\documentclass[10pt,journal,compsoc]{IEEEtran}
\usepackage{cite}
\usepackage{amsmath,graphicx}
\usepackage{epstopdf}
\usepackage{ragged2e}
\usepackage{xcolor}
\usepackage{amssymb}
\usepackage{pifont}
\usepackage{array}
\usepackage{booktabs}
\usepackage{bm}
\usepackage{multirow}
\usepackage{graphicx}
\usepackage{color}
\usepackage{float}
\usepackage{stfloats}
\usepackage[misc]{ifsym}
\usepackage{stmaryrd}
\usepackage{grffile}
\usepackage{subfig}
\usepackage{makecell}
\usepackage{threeparttable}
\usepackage{enumitem}
\usepackage{enumerate}
\usepackage[pagebackref=true,breaklinks=true,letterpaper=true,colorlinks,bookmarks=false]{hyperref}
\usepackage{colortbl}
\definecolor{shadow}{cmyk}{.08,0,0,0}
\usepackage{tikz}
\newcommand*{\circled}[1]{\lower.6ex\hbox{\tikz\draw (0pt, 0pt)%
    circle (.4em) node {\makebox[1em][c]{\tiny #1}};}}
\usepackage{diagbox}
\begin{document}

\title{Triple Spectral Fusion for \\ Sensor-based Human Activity Recognition}

\author{Ye Zhang, Longguang Wang, Qing Gao, Chaocan Xiang, Mohammed Bennamoun, and Yulan Guo, \IEEEmembership{Senior Member, IEEE}
\vspace{-0.1cm}
\IEEEcompsocitemizethanks{
\IEEEcompsocthanksitem
Ye Zhang, Qing Gao, and Yulan Guo are with the School of Electronics and Communication Engineering, the Shenzhen Campus of Sun Yat-sen University, Sun Yat-sen University, Shenzhen 518107, China. Longguang Wang is with the Aviation University of Air Force, Changchun 130012, China. Chaocan Xiang is with the College of Computer Science, Chongqing University, Chongqing 400044, China. Mohammed Bennamoun is with the Department of Computer Science and Software Engineering, the University of Western Australia, Perth, Western Australia 6009, Australia. Emails: \{zhangy2658, gaoqing2, guoyulan\}@mail.sysu.edu.cn, wanglongguang15@nudt.edu.cn, xiangchaocan@cqu.edu.cn, mohammed.bennamoun@uwa.edu.au.
}
\IEEEcompsocitemizethanks{
\IEEEcompsocthanksitem
This work was partially supported by the National Natural Science Foundation of China (No. 62372491, 62572082), the Aeronautical Science Fund of China (2025L0150M1001), the Guangdong Basic and Applied Basic Research Foundation (2023B1515120087, 2025A1515011954, 2026A1515012484), the Science and Technology Planning Project of Key Laboratory of Advanced IntelliSense Technology, Guangdong Science and Technology Department (2023B1212060024), the Shenzhen Key Laboratory of Navigation and Communication Integration (ZDSYS20210623091807023), Shenzhen Science and Technology Program (ZDCY20250901100201002), the Natural Science Foundation of Chongqing (No.CSTB2024NSCQ-JQX0020), and the Australian Research Council grant (DP260101891). Corresponding author: Yulan Guo.
}
}

\markboth{}%
{Shell \MakeLowercase{\textit{et al.}}: Bare Advanced Demo of IEEEtran.cls for IEEE Computer Society Journals}

\IEEEtitleabstractindextext{
\begin{abstract}
\justifying
The field of sensor-based human activity recognition (HAR) mainly uses posture, motion and context data of Inertial Measurement Units (IMUs) to identify daily activities. Despite the advancements in learning-based methods, it is challenging to perform information fusion from the temporal perspective due to the complexities in fusing heterogeneous sensor data and establishing long-term context correlations. This paper proposes a novel triple spectral fusion framework tailored for HAR. First, we develop an adaptive complementary filtering technique for noise suppression and organize each IMU's sensors into posture and motion modality nodes. Given that IMU nodes form a dynamic heterogeneous graph, we then apply adaptive filtering within the graph Fourier domain to merge both homogeneous and heterogeneous node information. Furthermore, an adaptive wavelet frequency selection approach is implemented to suppress context redundancy and shorten the length of features. This approach enhances both timestamp-based graph aggregation and the correlation of long-term contexts. Our framework uses adaptive filtering in the Fourier, graph Fourier, and wavelet domains, enabling effective multi-sensor fusion and context correlation. Extensive experiments on ten benchmark datasets demonstrate the superior performance of our framework. Project page: \url{https://github.com/crocodilegogogo/TSF-TPAMI2026}.
\end{abstract}
}
\IEEEaftertitletext{\vspace{-1.6\baselineskip}}
\maketitle

\IEEEdisplaynontitleabstractindextext

\IEEEpeerreviewmaketitle

\section{Introduction}\label{sec:introduction}

\IEEEPARstart{S}{ensor}-based human activity recognition (HAR) plays a key role in numerous fields, including health-care \cite{ricotti2023wearable,phan2022xsleepnet}, sport tracking \cite{dai2024hisc4d,nguyen2025class}, smart homes \cite{yang2024cross,thukral2025layout}, and intelligent manufacturing \cite{li2025stade}. HAR systems use signals from internal sensors, primarily Inertial Measurement Units (IMUs) in smartphones and wearable devices, to automatically identify human activities in daily life \cite{busso2025diversityone}. The typical process in HAR involves segmenting these sensor signals into equal-length subsequences, which are then classified into specific pre-defined activities \cite{haresamudram2025past}. Although vision-based methods enable non-contact activity monitoring and offer better data interpretability, sensor-based HAR has recently attracted increasing attention due to its advantages in unconstrained working conditions, enhanced privacy protection, and lower computational demands.

Dealing with data from multiple sensors over a time period, HAR faces two major challenges \cite{xie2025decomposing}: multi-sensor fusion and the association of activity context. Recent learning-based studies have extensively explored solutions to these issues \cite{chen2021deep}. Earlier works used shared or multi-branch convolutional filters for fusing data from multiple sensors, employing Convolutional Neural Networks (CNNs) to capture temporal information \cite{yang2015deep,laput2019sensing}. However, CNNs are somewhat limited in establishing long-term temporal associations effectively. To address this, hybrid models combining convolutional filters with Recurrent Neural Networks (RNNs) emerged \cite{ordonez2016deep,yao2017deepsense,ma2019attnsense}. RNNs are more adept than CNNs at extracting temporal context features. Building upon this, Graph Neural Networks (GNNs) were introduced to establish inherent sensor correlations, replacing convolutional filters \cite{liu2020globalfusion,abedin2021attend,miao2022towards}. More recently, to overcome the `forgetting' problem inherent in RNNs, Transformers have been incorporated into hybrid models, enabling the capture of long-term temporal dependencies \cite{mahmud2020human,li2021two,zhang2022if}. While these hybrid models have shown impressive results, they primarily focus on the temporal aspect of HAR, leaving several important challenges unaddressed.

\textbf{First}, current HAR methods do not fully utilize the distinct physical roles of different IMU sensors, leading to inadequate intra-IMU fusion and noise reduction. Many recent sensor fusion methods treat all IMU sensors uniformly \cite{laput2019sensing,ma2019attnsense,miao2022towards}. However, gravimeters and gyroscopes primarily record the posture state of the sensor carrier and are susceptible to high-frequency and low-frequency noise, respectively \cite{mahony2008nonlinear}. Additionally, linear accelerometers are used to capture motion information. This specific knowledge has not been thoroughly explored by previous methods. To address this, we propose an IMU fusion block with separate posture and motion branches. The posture branch includes an adaptive complementary filtering mechanism. Here, gravimeter and gyroscope features are independently processed through complementary low-pass and high-pass filters with an adjustable cut-off frequency. This approach reduces noise interference and leverages the frequency complementarity of gravimeters and gyroscopes for enhanced posture information. The motion branch, functioning separately from the posture branch, extracts motion information of the sensor carrier from the linear accelerometer features.

\textbf{Second}, the extraction and fusion of heterogeneous modality information have been underemphasized. Existing methods typically take sensors as graph nodes and rely on homogeneous GNNs (e.g., GAT \cite{abedin2021attend} and GCN \cite{miao2022towards}) to capture dependencies among sensors placed at multiple body positions. While effective in establishing spatial correlations, these methods often overlook heterogeneous node information \cite{liu2020globalfusion}, which is vital to capture individualized features arising from varied sensor types and positions. To address this, we aim to incorporate heterogeneous characteristics with homogeneous information. Extracting heterogeneous information directly in the time domain is challenging, we hence propose a modality node fusion block using an adaptive filtering mechanism in the graph Fourier domain. In our approach, a fully-connected graph is created, treating the posture and motion branches of different IMU blocks as nodes. High-pass and low-pass filters in the graph Fourier domain are then developed to extract heterogeneous and homogeneous node information, respectively. Following this, these filters are adaptively weighted to effectively fuse both homogeneous and heterogeneous information.

\textbf{Finally}, the issue of redundant temporal information presents significant challenges in establishing context correlations. The frequency of human activities is typically much lower than the sensor sampling rates \cite{van2013separating}, and sensors often capture meaningless activities. As a result, recorded data tends to contain a high level of redundancy, which impedes the effectiveness of temporal associations in existing RNN \cite{ordonez2016deep,yao2017deepsense,ma2019attnsense} and Transformer \cite{mahmud2020human,li2021two,zhang2022if} models and leads to increased computational demands. To address this, we have developed a temporal information fusion block that incorporates an adaptive wavelet time-frequency downsampling mechanism. Prior to each graph aggregation or context correlation process, discrete wavelet transform (DWT) is applied to decompose input features into high- and low-frequency components, each downsampled to half the original length. Only the primary component is adaptively chosen for further graph or temporal correlation. Through multiple decomposition and selection steps, we progressively map the activity sequence to its optimal time-frequency band, thereby diminishing redundancy and achieving efficient context correlation.

By synergistically integrating these blocks, we propose the \textbf{T}riple \textbf{S}pectral \textbf{F}usion (TSF) framework. TSF offers a comprehensive spectral perspective for HAR by operating across the Fourier, graph Fourier, and wavelet domains, allowing for a more effective fusion of information from multiple sensors and activity context. This paper extends our earlier conference version \cite{zhang2022if}, where we introduced the IF-ConvTransformer framework for HAR. The additional contributions of this paper, compared to our preliminary version, are as follows:

\begin{itemize}
	\item We introduce a novel modality node fusion block using an adaptive filtering mechanism in the graph Fourier domain. This block effectively integrates homogeneous and heterogeneous modality information, thereby improving multi-sensor fusion beyond what was achieved with the IF-ConvTransformer.

    \item A novel temporal information fusion block, incorporating an adaptive wavelet frequency selection mechanism, has been proposed. This block efficiently reduces temporal redundancies, enhancing context correlations and computational efficiency compared to the IF-ConvTransformer.

    \item We propose TSF, a novel HAR framework that employs specific filtering mechanisms across three spectral domains. The TSF framework demonstrates state-of-the-art performance on ten public HAR datasets.

\end{itemize}

\section{Related Work}\label{sec:RelatedWork}

This paper focuses on HAR methods, specifically addressing the challenges of multi-sensor fusion and context association, both of which are integral to the scope of this work.

\subsection{Multi-sensor Fusion}\label{sec:ms-fusion}

HAR tasks typically rely on multiple IMU sensors to capture the posture and motion states of human body. Multi-sensor fusion thus forms a significant challenge in HAR \cite{xie2025decomposing}. Existing methods for multi-sensor fusion in HAR can be categorized into three main types: classifier ensemble, data-level fusion, and feature-level fusion.

Classifier ensemble-based methods achieve multi-sensor fusion by combining the recognition results from different sensors. Guo et al. \cite{guo2016wearable} implemented a Multi-Layered Perceptron (MLP) classifier for each sensor, subsequently weighting these classifiers adaptively for consolidated classification results. Khan et al. \cite{khan2017detecting} used multiple autoencoders' reconstruction errors for fall detection. Chai et al. \cite{chai2025iot} fused the outputs of multiple classifiers, each being trained on time- and frequency-domain statistical features extracted from distributed body sensors. \textit{These methods are highly adaptable to varying sensor numbers but are less effective in establishing sensor correlations.}

Data-level fusion methods amalgamate sensor data before processing by the recognition model. Ha et al. \cite{ha2015multi} formed a 2D matrix by vertically stacking signals from different sensors and applied a 2D CNN to extract sensor dependencies. Li et al. \cite{li2025gesture} transformed sensor signals into time-frequency diagrams and concatenated them horizontally. However, these methods may overlook correlations of distantly placed sensors due to the limited reach of convolutional kernels. Addressing this, Jiang et al. \cite{jiang2015human} devised a stacking strategy allowing adjacent placement of signals from each sensor. \textit{Data-level fusion methods are straightforward but may underperform due to limited capability in capturing intrinsic sensor correlations.}

Feature-level fusion methods integrate features extracted from different sensors for recognition. Early works used manual techniques for multi-sensor fusion, such as Bayesian co-boosting \cite{wu2014bayesian} and mutual information-based feature selection \cite{fish2012feature}. With the advent of deep learning, shared conv-filters for sensor fusion were introduced \cite{yang2015deep,ordonez2016deep}, allowing influence from multiple sensors simultaneously. Some works handle sensors independently and then combine their outputs. Yao et al. \cite{yao2017deepsense} developed a multi-branch model using spectrogram matrices of each sensor as inputs to different branches, which are then fused by a conv-layer. Ma et al. \cite{ma2019attnsense} expanded on this with a branch attention mechanism. Recently, Betancourt et al. \cite{betancourt2020self} employed a self-attention mechanism to detect internal sensor correlations, akin to a homogeneous oriented Graph Attention Network (GAT) \cite{wang2022survey}. Abedin et al. \cite{abedin2021attend} applied GAT for a cross-channel interaction encoder. Liu et al. \cite{liu2020globalfusion} proposed a hierarchical GAT-based framework to fuse spatial positions in shallower layers and sensor modalities in deeper layers. \textit{Feature-level fusion methods generally outperform others due to their effectiveness in capturing internal sensor correlations.}

Despite extensive exploration, multi-sensor fusion in HAR still faces challenges, as discussed in Section~\ref{sec:introduction}. \textit{Current methods often treat different IMU sensors equally, not fully considering their distinct roles and physical characteristics. Additionally, the heterogeneous nature of feature extraction across various sensors and body positions is frequently underestimated.}

\subsection{Temporal Information Fusion}\label{sec:ti-fusion}

Human activities consist of various basic movements that span from a few seconds to several minutes in duration. This variability makes it crucial to efficiently fuse temporal information, as indicated in references \cite{ye2018learning,zhang2022multi}.

Early studies adhered to the conventional paradigm, creating manually-designed features to capture context correlations. Parkka et al. \cite{parkka2006activity} extracted features such as mean, variance, and kurtosis of signals, and used the decision trees for recognition tasks. Building on this, Altun et al. \cite{altun2010comparative} employed principal component analysis to streamline feature channels. Preece et al. \cite{preece2008comparison} evaluated a variety of features in both time-domain and frequency-domain, revealing the superior efficacy of frequency-based features. Ward et al. \cite{ward2006activity} designed a linear discriminate analysis method to identify temporal dependencies, followed by the use of HMMs for classification. \textit{Despite their proven effectiveness, these traditional techniques are typically labor-intensive and heavily dependent on specialized expertise.}

Deep learning techniques have gained popularity for analyzing temporal correlations \cite{guo2026deep}. They are generally classified into three groups: CNN-based, RNN-based, and Transformer-based approaches. Pioneering works \cite{yang2015deep,laput2019sensing} used CNNs to extract context features from both sensor signals and time-frequency transformed data. However, CNNs have limitations in capturing long-term correlations \cite{chen2021deep}. To address this, RNN models like Long Short-Term Memory (LSTM) and Gated Recurrent Unit (GRU) were introduced. For instance, Ord{\'o}{\~n}ez et al. \cite{ordonez2016deep} combined CNNs with LSTMs. Following this, Abedin et al. \cite{abedin2021attend} added a temporal attention mechanism to emphasize more critical timestamps. Addressing the `forgetting' issue in RNN models, Yao et al. \cite{yao2017deepsense} and Liu et al. \cite{liu2020globalfusion} used the Short-Time Fourier Transform (STFT) to shorten input data lengths, applying ConvGRUs for global context correlation encoding. However, STFT might result in the loss of context information. More recently, Transformers, superior in modeling long-term time series data correlations \cite{li2023difFormer,Eldele2023self}, have become a favored choice. Mahmud et al. \cite{mahmud2020human} proposed the first Transformer model for HAR, which included multiple self-attention blocks and a global temporal attention block. Jeong et al. \cite{jeong2024precyse} applied patch embedding to incorporate local context prior to the global Transformer stage. Additionally, Shao \cite{shao2023convboost} and Wang \cite{wang2024multi} combined CNNs with Transformers to capture both local and global context features.

Numerous studies have explored temporal information fusion, constructing both local and global temporal correlations. \textit{However, as mentioned in Section~\ref{sec:introduction}, current methods struggle with effectively managing context redundancy arising from the disparity between high sensor sampling rates and low activity frequencies. These redundancies complicate temporal correlations and increase computational demands.}

\section{Method}\label{sec:4}

\subsection{Overview}
\label{sec:4.1}

\begin{figure*}[t]
\centering
\footnotesize
\begin{tabular}{c}

\includegraphics[width=16.6cm]{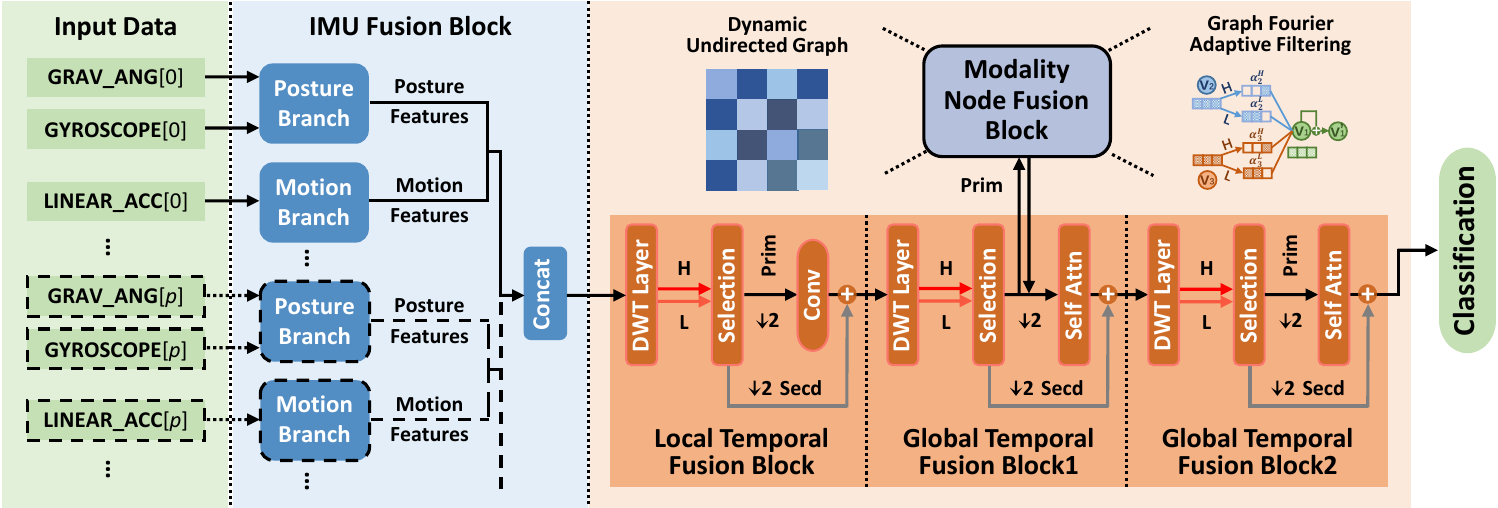}\\

\end{tabular}

\caption{Overall TSF framework depicting the IMU fusion block, the modality node fusion block and the temporal information fusion block, ending in classification. Here, $p$ denotes the index of IMUs, with $\textbf{GRAV\underline{\hspace{0.5em}}\hspace{0.15em}ANG}[p]$, $\textbf{GYROSCOPE}[p]$, and $\textbf{LINEAR\underline{\hspace{0.5em}}\hspace{0.15em}ACC}[p]$ representing the gravimeter posture angle, gyroscope angular velocity, and linear acceleration of the $p$-th IMU. The dotted boxes represent the parts that need to be added for wearable device-based HAR tasks. These parts involve additional posture and motion branches. Furthermore, L and H denote the low- and high-frequency components obtained by DWT, while Prim and Secd indicate the selected primary component and the unselected secondary component, respectively, each being down-sampled by a factor of two (denoted as $\downarrow$2). Self Attn refers to the self-attention layer.}
\label{fig1}

\end{figure*}

In HAR tasks, the smartphones and wearable devices are often placed at specific body locations, incorporating one or several IMUs. An IMU commonly comprises two built-in sensors: an accelerometer and a gyroscope. The accelerometer can further be separated as a gravimeter and a linear accelerometer \cite{van2013separating}. Among them, the gravimeter and gyroscope function to determine the posture angles of the wearer, while the linear accelerometer tracks the wearer's motion. These IMU sensors capture triaxial signals that reflect the body's state, resulting in the collected data forming a multi-channel time series \textbf{\emph{S}}.
\begin{equation} \label{e3.1}
\textbf{\emph{S}} = [\textbf{\emph{s}}_1,...,\textbf{\emph{s}}_t,...],
\end{equation}
where $\textbf{\emph{s}}_t = [\textbf{\emph{s}}_{t11},...,\textbf{\emph{s}}_{tpk},...\textbf{\emph{s}}_{tPK}]^T$, and $\textbf{\emph{s}}_{tpk}$ is the triaxial data of the $k$-th sensor placed at the $p$-th body location at the $t$-th timestamp. $P$ is the number of body locations carrying IMUs, and $K$ is the number of different sensors in an IMU. Given a recorded multi-dimensional sequence \textbf{\emph{S}}, it is first segmented into shorter fragments. Then, each fragment is mapped into a predefined activity label.

Recognizing an activity from multi-sensor signals poses two major challenges: multi-sensor fusion and context association. Existing learning-based methods, such as those referenced in \cite{chen2021deep}, usually address these challenges from a temporal perspective. However, they struggle with extracting heterogeneous sensor information and mitigating redundancy and noise. These issues are more manageably addressed in the spectral domain. Despite the introduction of spectral transformation to HAR by methods like those in \cite{yao2017deepsense, liu2020globalfusion}, most such transformations are performed at the data level, resulting in unsatisfactory performance due to a lack of adaptation ability.

To address the HAR task challenges, we propose our TSF framework, extending spectral transformations to the feature level and applying spectral filtering on deep features for information fusion. In this way, our TSF framework effectively adapts to data characteristics through end-to-end optimization. As depicted in Figure~\ref{fig1}, TSF consists of three spectral fusion blocks: the IMU fusion block integrates sensors of each IMU into posture and motion nodes using adaptive complementary filtering (Section~\ref{sec:4.2}); the modality node fusion block merges homogeneous and heterogeneous node information in the graph Fourier domain (Section~\ref{sec:4.3}); and the temporal information fusion block applies adaptive wavelet frequency selection to suppress temporal redundancies, significantly reducing feature length and facilitating timestamp-wise node aggregation and long-term context correlation (Section~\ref{sec:4.4}). Post-processing, these features are fed into an average pooling layer and a linear layer for classification.

\subsection{IMU Fusion Block via Adaptive Complementary Filtering}
\label{sec:4.2}

Section~\ref{sec:introduction} highlights that different IMU sensors perform distinct physical roles. Despite this, recent sensor fusion methods often overlook this specificity, treating various IMU sensors uniformly \cite{laput2019sensing,ma2019attnsense,miao2022towards}, which results in inadequate intra-IMU fusion. To handle this issue, we have designed an IMU fusion block that utilizes adaptive complementary filtering. Next, we first review the complementary filter approach (Section~\ref{sec:4.2.1}), and then elaborate on how our fusion block integrates sensor data while preserving the unique characteristics of each sensor (Section~\ref{sec:4.2.2}).

\subsubsection{Rethinking the Complementary Filter}
\label{sec:4.2.1}

Gravimeters and gyroscopes are typically used to measure the posture states of sensor carriers. However, gravimeter data is prone to high-frequency noise interference, while gyroscope data suffers from low-frequency noise, leading to a significant `drift' issue \cite{jung2007inertial}. To improve posture estimation, many unmanned aerial vehicle systems adopt a complementary filter to integrate data from these sensors \cite{mahony2008nonlinear}. This filter processes gravimeter data through low-pass filters and gyroscope data through high-pass filters. The combined outputs of these filters then provide refined posture data. The complementary filter's function is governed by a discrete-frequency equation, defined by:
\begin{equation} \label{e4.1}
\begin{aligned}
\textbf{att}(z) = {}& \textbf{grav\underline{\hspace{0.5em}}\hspace{0.15em}ang}(z) \cdot \frac{1}{\tau \cdot (1-z^{-1})/T + 1}\\
&{} + \textbf{gyro}(z) \cdot \frac{\tau}{\tau \cdot (1-z^{-1})/T + 1},
\end{aligned}
\end{equation}
where $\textbf{att}(z)$, $\textbf{grav\underline{\hspace{0.5em}}\hspace{0.15em}ang}(z)$, and $\textbf{gyro}(z)$ represent the Z-transformations of estimated posture angle, gravimeter posture angle, and gyroscope angular velocity, respectively. The parameter $\tau$ is a predefined constant in this context, where $1/\tau$ defines the cut-off frequency. Additionally, $T$ is the sampling interval.

Equation~\ref{e4.1} can be transformed to the discrete-time form using the inverse Z-transform:
\begin{equation} \label{e4.2}
\begin{aligned}
\textbf{att}(t) = {}& \alpha \cdot (\textbf{att}(t-1) + \textbf{gyro}(t) \cdot T)\\
& + (1-\alpha) \cdot \textbf{grav\underline{\hspace{0.5em}}\hspace{0.15em}ang}(t),
\end{aligned}
\end{equation}
where $t$ is the timestamp, $\alpha$ is $\tau/(\tau+T)$.

As a recursive equation, Equation~\ref{e4.2} can be further expanded as:
\begin{equation} \label{e4.3}
\begin{aligned}
\textbf{att}(t) = {}& \alpha^t \cdot \textbf{grav\underline{\hspace{0.5em}}\hspace{0.15em}ang}(0) {} \\
&{} + \alpha^{t-1} \cdot ( (1-\alpha) \textbf{grav\underline{\hspace{0.5em}}\hspace{0.15em}ang}(1) + \alpha \cdot \textbf{gyro}(1) \cdot T ) {} \\
&{} + \alpha^{t-2} \cdot ( (1-\alpha) \cdot \textbf{grav\underline{\hspace{0.5em}}\hspace{0.15em}ang}(2) + \alpha \cdot \textbf{gyro}(2) \cdot T ) {} \\
&{} + ... {} \\
& + ( (1-\alpha) \cdot \textbf{grav\underline{\hspace{0.5em}}\hspace{0.15em}ang}(t) + \alpha \cdot \textbf{gyro}(t) \cdot T ).
\end{aligned}
\end{equation}
\vspace{0cm}
\vspace{0cm}

In Equation~\ref{e4.3}, two key observations are noted:

1) The posture state at timestamp $t$ is influenced by all previous timestamps, but their contributions diminish exponentially over time, as governed by weighting coefficients ($\alpha^{t-i}$, where $i$ indicates the $i$-th timestamp).

2) For each timestamp except for the initial one, the weighting coefficients of the gravimeter angle (1-$\alpha$) and the integral gyroscope readings ($\alpha$) always sum up to 1.

Despite its widespread success, the complementary filter has shortcomings. \textbf{First}, long-duration integration of gyroscope readings can accumulate small biases, leading to a serious drift issue \cite{jung2007inertial}. This problem can be mitigated by discarding earlier timestamps and focusing on more recent ones, a strategy inspired by the first observation. To this end, we propose convolutional kernels as cut-off windows to discard stale information while preserve latest cues.

\textbf{Second}, determining the optimal value of $\alpha$ to balance the gravimeter and gyroscope data is challenging due to real-world noise complexities \cite{mahony2008nonlinear}. Based on the second observation, the proposed solution uses an attention mechanism to automatically learn the value $\alpha$ at each timestamp.

\textbf{In conclusion, our aim is to redevelop the complementary filter into a trainable block, effectively addressing its inherent flaws.}

\subsubsection{Architecture of the IMU Fusion Block}
\label{sec:4.2.2}

\begin{figure}[t]
\centering
\footnotesize
\begin{tabular}{c}

\includegraphics[width=8.66cm]{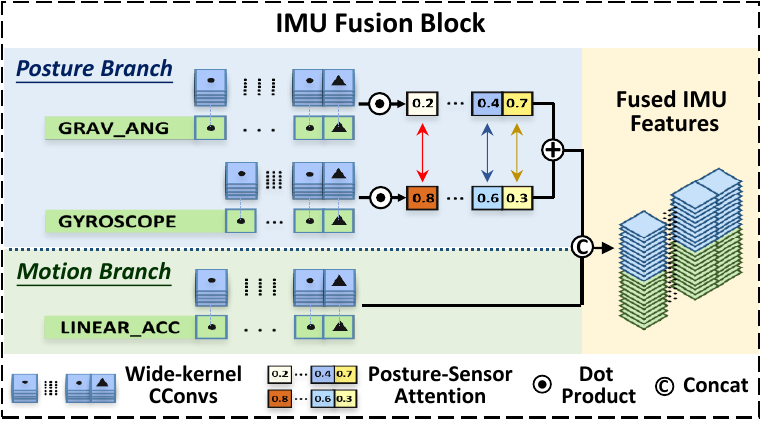}\\

\end{tabular}
\caption{Block diagram of the IMU fusion block within the TSF framework, demonstrating the separate processing branches of posture and motion sensor data and their concatenation into fused IMU features. In particular, the posture branch integrates the gravimeter and gyroscope data using causal convolutions (CConvs) \cite{Bai2018} along with a sensor attention mechanism. `$\blacktriangle$' in the kernels denotes the feature aggregation position of the CConvs.}
\label{fig2}
\end{figure}

The IMU fusion block in our design caters to two types of sensor data from IMUs: gravimeters and gyroscopes to measure the posture state, and linear accelerometers to measure the motion state of sensor carriers. Accordingly, the IMU fusion block is structured into two branches: the posture branch and the motion branch, as depicted in Figure~\ref{fig2}.

It is important to note that if the gravimeter and linear acceleration signals are not explicitly available, they can be derived from the acceleration signals using a Butterworth filter \cite{selesnick1998generalized}. We have selected a cutoff frequency of 0.3 Hz for this filter, based on the assumption that most daily human movements occur at frequencies above this threshold \cite{van2013separating}.

\textbf{(1) The Posture Branch}

Drawing inspiration from the complementary filter, the posture branch is designed to fuse features from gravimeter and gyroscope data into cohesive posture features. This branch comprises two main elements: wide-kernel causal convolutions (CConvs) \cite{Bai2018} and the posture-sensor attention mechanism, both motivated by observations made in Section~\ref{sec:4.2.1} of the complementary filter.

\textbf{Wide-kernel CConvs:} To address the first observation that the current posture state is more strongly correlated with recent timestamps, we use a 1D convolutional kernel to act as a sliding cut-off window. This window retains sensor readings from nearby timestamps, while discarding those from more distant ones.

However, standard convolutions inadvertently incorporate future information, which violate the complementary filter principle that relies solely on past and current data for posture estimation. To resolve this, we employ wide-kernel CConvs, as illustrated in Figure~\ref{fig2}. Unlike typical convolutions, CConvs \cite{Bai2018} accumulate features at the kernel's last position rather than its center, effectively preventing future information leakage. Furthermore, compared to the usual $1\times3$ or $1\times5$ kernels used in HAR \cite{ma2019attnsense}, wide kernels ($1\times11$ in this case) offer richer historical information through broader coverage. These kernels are applied separately to gravimeter and gyroscope signals, with the resultant features fused for output. To align with Equation~\ref{e4.3}, the CConv kernels for gravimeter data ($1\times11$) include one additional point over the gyroscope kernels ($1\times10$). This extra point serves to weight the initial posture state at timestamp `0'.

\textbf{Posture-sensor Attention Mechanism:} As discussed in Section~\ref{sec:4.2.1}, the weighting coefficients of gravimeter and gyroscope data are vital, as they determine the cut-off frequencies of the complementary filter. Building on the second observation, where these coefficients sum up to 1, we use the attention mechanism to adaptively balance the importance of gravimeter and gyroscope, as shown in the posture branch of Figure~\ref{fig2}. Specifically, the attention weight for each sensor at each timestamp $\mu_{tk}$ is calculated using Equation~\ref{e4.4}, normalized through a softmax function to obtain $attn_{tk}$ (see Equation~\ref{e4.5}), and then applied to the posture sensor features $\textbf{\emph{v}}_{tk}$ (see Equation~\ref{e4.6}):
\begin{eqnarray} \label{e4.4}
\mu_{tk} = {\rm tanh}(\textbf{\emph{W}}_0 \cdot \textbf{\emph{v}}_{tk} + \textbf{\emph{b}}_0),
\end{eqnarray}
\begin{eqnarray} \label{e4.5}
attn_{tk} = \frac{{\rm exp}(\mu_{tk})}{\sum_k {\rm exp}(\mu_{tk})},
\end{eqnarray}
\begin{eqnarray} \label{e4.6}
O_{tk} = attn_{tk} \cdot \textbf{\emph{v}}_{tk},
\end{eqnarray}
where $\textbf{\emph{W}}_0$ and $\textbf{\emph{b}}_0$ are parameters of a single-layer MLP, $attn_{tk}$ represents the weighting coefficient of the $k$-th posture sensor at each timestamp $t$, and $O_{tk}$ the output feature.

\textbf{(2) The Motion Branch}

The motion branch is dedicated to capture the motion state of the sensor carrier, primarily processing linear accelerometer data to extract relevant motion features. As shown in Figure~\ref{fig2}, wide-kernel CConvs are used in this branch, with kernel sizes matching those applied to the gravimeter data. This setup ensures that the integration of linear acceleration data over the CConv kernels accurately reflects the velocity changes experienced by sensor carrier.

\subsection{Modality Node Fusion Block via Adaptive Filtering in Graph Fourier Domain}
\label{sec:4.3}

The IMU fusion block integrates each IMU's sensors into posture and motion modality branches. However, effectively fusing these branches poses a challenge due to their differing physical meanings and spatial variances. While existing methods model these relationships via spatial graphs, their time-domain GNNs \cite{abedin2021attend,miao2022towards} often struggle to capture the heterogeneous node information. To address this, we propose a modality node fusion block that employs adaptive filtering in the graph Fourier domain, enabling integration of both homogeneous and heterogeneous information. We first describe the process to transform the posture and motion branches into a spatial graph and the graph Fourier domain (Section~\ref{sec:4.3.1}), followed by the design of our low-pass and high-pass filters and their adaptive fusion mechanism (Section~\ref{sec:4.3.2}).

\subsubsection{Graph Fourier Transform}
\label{sec:4.3.1}

Taking each posture and motion branch as a node, we establish a fully connected undirected graph, represented as $\textbf{\emph{G}} = (\textbf{\emph{V}},\textbf{\emph{E}})$, where \textbf{\emph{V}} is the set of graph nodes, \textbf{\emph{E}} is the set of edges, and \emph{N} is the number of nodes in $\textbf{\emph{G}}$. The normalized Laplacian matrix \textbf{\emph{L}} of \textbf{\emph{G}} is defined as $\textbf{\emph{L}} = \textbf{\emph{I}}-\textbf{\emph{D}}^{-\frac{1}{2}}\textbf{\emph{A}}\textbf{\emph{D}}^{-\frac{1}{2}}$, with \textbf{\emph{I}} being the identity matrix, $\textbf{\emph{A}}$ the adjacency matrix (edge weights) of $\textbf{\emph{G}}$, and \textbf{\emph{D}} the diagonal degree matrix. Since \textbf{\emph{L}} is a real symmetric matrix, it has a complete set of orthonormal eigenvectors ${\{\bm{\emph{$\mu$}}_l\}}_{l=1}^{N} \in \mathbb{R}^{N}$ with eigenvalues $\lambda_l \in [0,2]$ \cite{chung1997spectral}. Hence, \textbf{\emph{L}} can be decomposed as:
\begin{equation} \label{e4.7}
\textbf{\emph{L}} = \bm{\mathit{U}}\bm{\emph{$\Lambda$}}\bm{\mathit{U}}^{\top},
\end{equation}
where $\bm{\mathit{U}}$ are the eigenvectors, and $\bm{\emph{$\Lambda$}}$ is the diagonal eigenvalue matrix.

With $\bm{\mathit{U}}$ as mapping bases, the Graph Fourier Transform (GFT) and inverse GFT of input node features $\textbf{\emph{X}}$ are defined as $\textbf{\emph{\^{X}}} = \bm{\mathit{U}}^{\top}\textbf{\emph{X}}$ and $\textbf{\emph{X}} = \bm{\mathit{U}}\textbf{\emph{\^{X}}}$, respectively. A filter $\emph{h}$ can be applied to the eigenvalues $\lambda_l$ of $\bm{\emph{$\Lambda$}}$ to modulate frequencies,
\begin{equation} \label{e4.8}
\begin{aligned}
\textbf{\emph{Y}} = {}& \bm{\mathit{U}}\bm{\emph{h($\Lambda$)}}\bm{\mathit{U}}^{\top}\textbf{\emph{X}}\\
 = {}& \bm{\mathit{U}}{\rm diag}([h(\lambda_1),h(\lambda_2),\cdot\cdot\cdot,h(\lambda_N)])\bm{\mathit{U}}^{\top}\textbf{\emph{X}}.
\end{aligned}
\end{equation}

For instance, GCN \cite{kipf2016semi} performs filtering through $\textbf{\emph{Y}} = (2\textbf{\emph{I}}-\textbf{\emph{L}})\textbf{\emph{X}}$. Given that the original GFT can be considered as a linear high-pass filter, GCN acts as a linear low-pass filter.

\subsubsection{Adaptive Filtering in Graph Fourier Domain}
\label{sec:4.3.2}

\begin{figure}[t]
\centering
\footnotesize
\begin{tabular}{c}

\includegraphics[width=8.66cm]{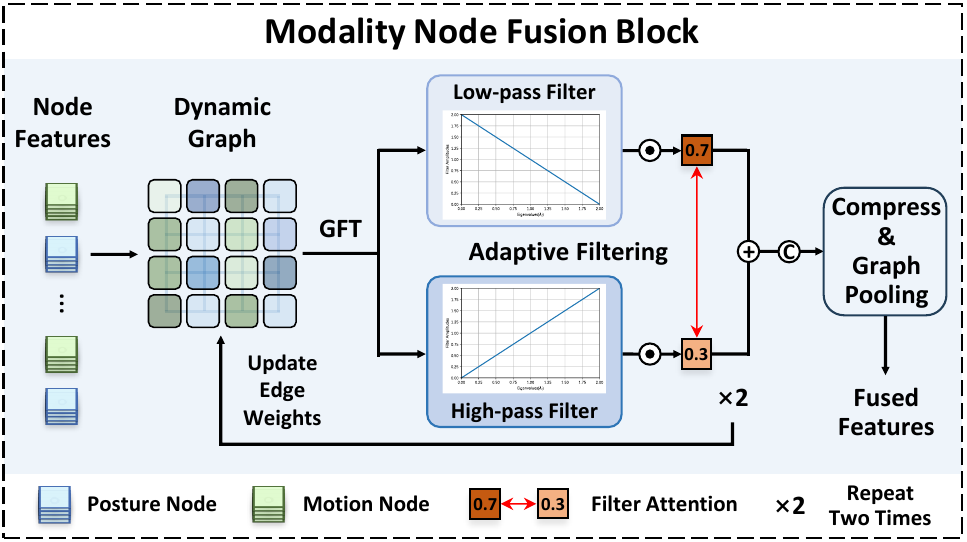}\\

\end{tabular}
\caption{Workflow of the modality node fusion block within the TSF framework, including the dynamic graph updating and the adaptive filtering process. The adaptive filtering employs high-pass and low-pass filters integrated via an attention mechanism. The final graph-level fusion is achieved by concatenating input features with filtering outputs, followed by channel compression and average graph pooling.}
\label{fig3}
\end{figure}

As shown in Figure~\ref{fig3}, adaptive filtering in the graph Fourier domain involves a high-pass filter to capture the heterogeneous information by retaining node differences, and a low-pass filter to smooth node features and extract homogeneous information. An attention mechanism is used to integrate these filters, adapting the balance between heterogeneous and homogeneous information. Additionally, to accommodate varying node correlations, the adjacency matrix and spectral mapping bases are dynamically updated. This ensures that the underlying GFT basis is also adaptively adjusted to match the evolving graph structure.

Our approach includes two adaptive filtering layers. The input features are concatenated with the output of each filtering layer, compressed in the channel dimension, and then processed through an average pooling layer for comprehensive graph-level fusion.

\textbf{(1) Filter Design}

The low-pass filter ($\textbf{\emph{F}}_L$) is represented by GCN, while the original GFT serves as the high-pass filter ($\textbf{\emph{F}}_H$):
\begin{equation} \label{e4.9}
\left\{
\begin{aligned}
\textbf{\emph{F}}_L & = \textbf{\emph{I}} + \textbf{\emph{D}}^{-\frac{1}{2}}\textbf{\emph{A}}\textbf{\emph{D}}^{-\frac{1}{2}} = 2\textbf{\emph{I}} - \textbf{\emph{L}}, \\
\textbf{\emph{F}}_H & = \textbf{\emph{I}} - \textbf{\emph{D}}^{-\frac{1}{2}}\textbf{\emph{A}}\textbf{\emph{D}}^{-\frac{1}{2}} = \textbf{\emph{L}}.
\end{aligned}
\right.
\end{equation}

Both filters are visualized in Figure \ref{fig3}. They are double applied to enhance the non-linearity of filtering. Using the filter attention mechanism, $\textbf{\emph{F}}_L$ and $\textbf{\emph{F}}_H$ can be integrated:
\begin{equation} \label{e4.10}
\tilde{\bm{Y}} = \alpha_L\textbf{\emph{F}}_L\textbf{\emph{X}} + \alpha_H\textbf{\emph{F}}_H\textbf{\emph{X}}.
\end{equation}
where $\alpha_L$ and $\alpha_H$ are the attention weights of different filters ($\alpha_L + \alpha_H = 1$). With Equation~\ref{e4.9}, Equation~\ref{e4.10} can be rewritten as:
\begin{equation} \label{e4.11}
\begin{aligned}
\tilde{\bm{Y}} & = \alpha_L(\textbf{\emph{I}} + \textbf{\emph{D}}^{-\frac{1}{2}}\textbf{\emph{A}}\textbf{\emph{D}}^{-\frac{1}{2}})\textbf{\emph{X}} + \alpha_H(\textbf{\emph{I}} - \textbf{\emph{D}}^{-\frac{1}{2}}\textbf{\emph{A}}\textbf{\emph{D}}^{-\frac{1}{2}})\textbf{\emph{X}} \\
& = \textbf{\emph{X}} + (\alpha_L-\alpha_H)\textbf{\emph{D}}^{-\frac{1}{2}}\textbf{\emph{A}}\textbf{\emph{D}}^{-\frac{1}{2}}\textbf{\emph{X}} \\
& = \textbf{\emph{X}} + \textbf{\emph{D}}^{-\frac{1}{2}}\tilde{\bm{A}}\textbf{\emph{D}}^{-\frac{1}{2}}\textbf{\emph{X}},
\end{aligned}
\end{equation}
where $(\alpha_L-\alpha_H) \in [-1,1]$. Since $\alpha_L$ and $\alpha_H$ represent the attention weights of $\textbf{\emph{F}}_L$ and $\textbf{\emph{F}}_H$, respectively, the filtering process is dominated by low-pass filters when $(\alpha_L-\alpha_H) > 0$, and by high-pass filters when $(\alpha_L-\alpha_H) < 0$. A value of $(\alpha_L-\alpha_H)$ near $0$ indicates that the neighboring nodes have less importance for the current node. Moreover, $(\alpha_L-\alpha_H)$ can be integrated into the adjacency matrix $\textbf{\emph{A}}$, transforming the problem into constructing a signed adjacency matrix $\tilde{\bm{A}}$.

\textbf{(2) Dynamic Adjacency Matrix Construction}

Due to the varying node correlations across different activities, subjects, and timestamps, constructing the signed adjacency matrix $\tilde{\bm{A}}$ rely solely on task-specific prior knowledge is challenging. To address this, $\tilde{\bm{A}}$ is learned automatically, initially set as a matrix of ones (representing a fully-connected graph). The edge weights of $\tilde{\bm{A}}$ are then updated based on node feature multiplication. Given the influence of $(\alpha_L-\alpha_H)$, $\tilde{\bm{A}}$ contains elements within the range $[-1,1]$, allowing it to adapt dynamically to the specific features of data. Therefore, elements in $\tilde{\bm{A}}$ are calculated as:
\begin{equation} \label{e4.12}
\alpha_{ij}^{G} = {\rm tanh}(\textbf{\emph{W}}_1[\textbf{\emph{x}}_i * \textbf{\emph{x}}_j] + \textbf{\emph{b}}_1).
\end{equation}

In this process, node features $\textbf{\emph{x}}_i$ and $\textbf{\emph{x}}_j$ are multiplied with parameters $\textbf{\emph{W}}_1$ and $\textbf{\emph{b}}_1$ from a two-layer MLP and a tanh activation function to calculate the elements of $\tilde{\bm{A}}$. These operations ensure $\tilde{\bm{A}}$'s symmetry, maintaining its Laplacian matrix form, which is crucial for the applicability and effectiveness of the designed filters. Note that, guided by the filter attention weights $(\alpha_L-\alpha_H)$, negative values in $\tilde{\bm{A}}$ reflect node dissimilarity and call for preserving heterogeneous information, whereas positive values reflect similarity and support retaining homogeneous information.

\subsection{Temporal Information Fusion Block via Adaptive Wavelet Frequency Selection}
\label{sec:4.4}

Long-term context information is essential in HAR. Recent methods, including Transformers \cite{zhang2022if,wang2024multi}, effectively capture these long-term dependencies but often suffer from high computational complexity. Since the frequencies of human activities are generally much lower than sensor sampling rates, there is a significant redundancy in context \cite{yadav2021review}. Thus, we construct a temporal information fusion block that utilizes DWT to downsample input features into half-length sub-bands, and then adaptively retains dominant frequency components, thereby reducing redundancy and computation. We introduce the DWT process in Section~\ref{sec:4.4.1} and the adaptive frequency selection mechanism in Section~\ref{sec:4.4.2}.

\subsubsection{Discrete Wavelet Transform}
\label{sec:4.4.1}

Wavelets serve as effective tools for time-frequency signal analysis in signal processing \cite{mallat1989theory}. Using DWT, an activity sequence $\textbf{\emph{s}} = \{s_t\}_{t=1}^{T_s}$ ($T_s$ being data length) can be split into low-frequency component $\textbf{\emph{s}}^L = \{s^{L}_{t_d}\}_{t_d=1}^{T_s/2}$ and high-frequency component $\textbf{\emph{s}}^H = \{s^{H}_{t_d}\}_{t_d=1}^{T_s/2}$ in half length of $T_s$:
\begin{equation} \label{e4.13}
\left\{
\begin{aligned}
s^{L}_{t_d} & = \sum_ws_{2t_d-w}l_w, \\
s^{H}_{t_d} & = \sum_ws_{2t_d-w}h_w.
\end{aligned}
\right.
\end{equation}

In this equation, $\{l_w\}_{w=1}^{W}$ and $\{h_w\}_{w=1}^{W}$ represent the low-pass and high-pass filters associated with a chosen wavelet basis, where $W$ is the filter length. The primary operations in DWT are wavelet filtering and down-sampling.

\subsubsection{Architecture of the Temporal Information Fusion Block}
\label{sec:4.4.2}

\begin{figure}[t]
\centering
\footnotesize
\begin{tabular}{c}

\includegraphics[width=6.6cm]{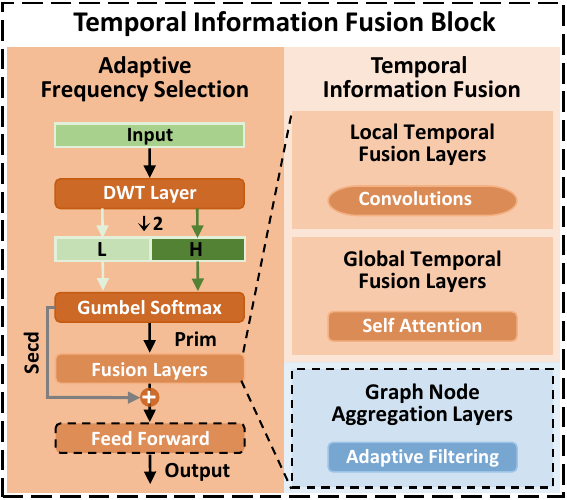}\\
\end{tabular}
\caption{Overview of the temporal information fusion block within the TSF framework, highlighting the adaptive frequency selection that precedes both local and global temporal fusion layers, as well as graph node aggregation layers. Notably, the temporal correlation and graph aggregation are performed exclusively on the primary wavelet frequency components, which are selected using the Gumbel softmax trick. After the global fusion layers, the feed-forward layers are applied to refine the learned representations.}

\label{fig4}
\end{figure}

As depicted in Figure~\ref{fig4}, the temporal information fusion block consists of adaptive wavelet frequency selection and temporal information fusion. The wavelet frequency selection part adaptively selects primary component from the decomposed low- and high-frequency features, using the secondary component as a residual term. Then, the temporal information fusion constructs context correlations from the selected primary component.

\textbf{(1) Adaptive Wavelet Frequency Selection}

This process filters out redundant context (like noise and meaningless activities) harmful to temporal information fusion \cite{yadav2021review}. Features are decomposed into two frequency components of halved length using DWT. A selection mechanism, aided by the Gumbel softmax distribution, identifies primary and secondary terms. Features undergo an average pooling layer, followed by an hourglass squeeze layer, then processed by a Gumbel softmax trick \cite{jang2016categorical,wang2023exploring} to create a softened binary mask. This mask helps distinguish between high- and low-frequency features, forming primary and secondary components. The primary component is used for constructing context correlations, while the secondary component is taken as a residual item to preserve information.

1) The input feature $\textbf{\emph{F}}^{in} \in \mathbb{R}^{2\times{C\times{N\times{L}}}}$ is fed into an average-pooling layer to produce $\textbf{\emph{F}}^{in} \in \mathbb{R}^{2\times{C}}$, followed by a squeeze layer to obtain $\textbf{\emph{F}}^s \in \mathbb{R}^{2\times1}$. The number `2' in the dimensions of $\textbf{\emph{F}}^{in}$ and $\textbf{\emph{F}}^s$ indicates that the high-frequency and low-frequency features are concatenated for selection. Here, $C$, $N$, and $L$ represent the feature channel number, node number, and temporal length of $\textbf{\emph{F}}^{in}$, respectively.

2) On top of $\textbf{\emph{F}}^s$, the Gumbel softmax trick is used to obtain a softened binary mask $\textbf{\emph{B}}^s$ as shown in Equation~\ref{e4.14}:
\begin{eqnarray} \label{e4.14}
\textbf{\emph{B}}^s[i] = \frac{{\rm exp}\left(\left(\textbf{\emph{F}}^s[i] + \textbf{\emph{G}}^s[i]\right)/\tau\right)}{\sum_{i=1}^{2}{\rm exp}\left((\textbf{\emph{F}}^s[i] + \textbf{\emph{G}}^s[i])/\tau\right)},
\end{eqnarray}
where $\textbf{\emph{G}}^s \in \mathbb{R}^{2}$ is a Gumbel noise tensor following Gumbel(0, 1) distribution, $\tau \in (0, \infty)$ is a temperature parameter. When $\tau \rightarrow \infty$, the elements in $\textbf{\emph{B}}^s$ become uniform at 0.5. Conversely, when $\tau \rightarrow 0$, the elements in $\textbf{\emph{B}}^s$ tend towards a one-hot distribution, rendering $\textbf{\emph{B}}^s$ binary. During training, $\tau$ starts high and gradually anneals to a lower value for optimization of the squeeze layer. In the inference phase, $\textbf{\emph{B}}^s$ becomes a one-hot distribution through the argmax operation, thus performing the selection between high-frequency and low-frequency features.

3) Using $\textbf{\emph{B}}^s$, the primary and secondary frequency components are obtained. Context correlations are constructed on the primary component. To prevent information loss, the secondary component is carried over as a residual item across the temporal fusion layers.

\textbf{(2) Temporal Information Fusion}

As depicted in Figure~\ref{fig4}, the layers of the temporal information fusion are applied solely on the selected primary wavelet frequency component, effectively halving the feature length before each context correlation. The fusion layers are categorized into local and global types based on their association scopes. Local fusion utilizes convolutional kernels to group isolated timestamps into semantically clear activity fragments \cite{zhang2022if}. Global fusion, on the other hand, employs self-attention layers to capture long-term dependencies and includes a feed-forward layer for feature projection. Position encoding \cite{vaswani2017attention} is applied prior to the initial self-attention layer to maintain temporal order.

\textbf{(3) Adaptive Wavelet Decomposition Routing}

In TSF, as shown in Figure~\ref{fig1}, the local fusion block precedes both the modality node fusion block and the global fusion blocks. The local fusion aggregates discrete temporal points into action segments and preliminarily compresses the temporal feature length, thereby enabling more efficient processing in subsequent fusion stages. Given the high computational demands of GNNs and self-attention layers within the fusion blocks, frequency selection occurs prior to graph aggregations and context associations to eliminate redundant computations, as illustrated in Figure~\ref{fig4}. Importantly, to prevent global temporal correlations from interfering with timestamp-wise graph aggregation, the modality node fusion block is situated before the first self-attention layer. Consequently, input features are significantly compressed to a quarter of their original length before graph aggregation and the initial global context fusion.

Viewing the temporal information fusion blocks through multi-level wavelet decomposition, they form a decomposition tree where each sample is directed to an appropriate wavelet frequency band via adaptive routing, ultimately reducing features to one-eighth of their original size.

\section{Experiments}\label{sec:Experiments}

\subsection{Experimental Setup}
\label{sec:5.1}

This section details the benchmark datasets, experimental settings, and evaluation metrics used in our analysis.

\subsubsection{Dataset Description}
\label{sec:5.1.1}

Ten public HAR datasets are used. They are categorized into smartphone-based and wearable device-based datasets:

\textbf{(1) Smartphone-based Datasets:} Five datasets recorded using smartphones with a single IMU are described in Table~\ref{tab1}, including participant counts, number of activities, sampling rates, and segmentation parameters (window sizes and sliding overlaps). For the HAPT \cite{reyes2016transition} and SHL2018 \cite{gjoreski2018university} datasets, data segmentation is pre-existing. The MotionSense \cite{malekzadeh2020privacy}, HHAR \cite{stisen2015smart}, and MobiAct \cite{chatzaki2016human} datasets follow the specific segmentation protocols outlined in \cite{malekzadeh2020privacy}, \cite{stisen2015smart}, and \cite{chatzaki2016human}, respectively. The HHAR dataset's varying sampling rates are unified using linear interpolation.

\textbf{(2) Wearable Device-based Datasets:} These five datasets described in Table~\ref{tab2}, are recorded using wearable devices with internal IMU sensors at multiple body locations. They contain more spatial information, leading to generally smaller segmentation window sizes. The DSADS dataset \cite{yao2018efficient} uses a preset 125-reading window without overlap, while the other four datasets (i.e., Oppo \cite{chavarriaga2013opportunity}, PAMAP2 \cite{reiss2012introducing}, RealWorld \cite{sztyler2016body}, and SHO \cite{shoaib2014fusion}) adopt a 48-reading window with 24 readings overlap as per \cite{abedin2021attend}. The Oppo dataset specifically uses data from IMUs on backs, arms, and shoes, covering seven body positions.

\subsubsection{Experimental Settings}
\label{sec:5.1.2}

\textbf{(1) Data Preprocessing}

For each dataset utilized, the triaxial signals from gravimeters, accelerometers, and gyroscopes are selected for HAR tasks. The signals of each sensor type are $z$-normalized individually. Subsequently, the normalized signals are segmented into smaller sections using a sliding window technique, with specific segmentation parameters detailed in Table~\ref{tab1} for smartphone-based datasets and Table~\ref{tab2} for wearable device-based datasets.

\textbf{(2) Implementation Details}

Implementation details of the TSF framework are provided in Table~\ref{tab3}. The model, built with Pytorch and run on a PC with two Nvidia RTX 2080Ti GPUs, is initialized with He's method \cite{he2015delving} and employs the Adam optimizer \cite{kingma2014adam}. A uniform training step count of 30 and a batch size of 128 are set for all datasets except SHL2018, which has 50 training steps, and Oppo and DSADS datasets, which have a batch size of 64. The learning rate starts at 0.0005 and is reduced by half every five epochs. Daubechies4 (DB4) wavelet basis is used, with the Gumbel softmax trick's temperature parameter $\tau$ starting at 1 and annealing to 0.5. Mixup data augmentation \cite{zhang2017mixup} is also utilized. The validation set, 10$\%$ of the training data, helps choose the model for performance evaluation based on the best validation accuracy.

Most comparison models are implemented using the authors' provided codes \cite{ordonez2016deep,yao2017deepsense,mahmud2020human,abedin2021attend,miao2022towards,yang2022predictive,zhang2022if,shao2023convboost}, with the remaining models reproduced based on their published descriptions \cite{liu2020globalfusion}. To ensure fairness, all models are adjusted to have a similar number of network parameters to the baseline method used for smartphone-based datasets. This adjustment is achieved by fine-tuning the number of channels without altering the overall model architecture. The parameter configurations are then consistently applied across wearable device-based datasets.

\begin{table}[t]
\renewcommand\arraystretch{1.2}
\centering

\scriptsize
\caption{Comparative overview of smartphone-based HAR datasets, detailing demographics, data acquisition settings, and segmentation parameters.}\label{tab1}
\vspace{-0.3cm}

\begin{tabular}{|c|c|c|c|c|c|}
\hline
\diagbox[innerwidth=2cm]{Attributes}{Datasets} & \makecell{HAPT} & \makecell{Motion-\\Sense} & \makecell{SHL2018} & \makecell{HHAR} & \makecell{MobiAct} \tabularnewline
\hline
Participant Count & 30 & 24 & 1 & 9 & 66\tabularnewline
\hline
Activity Number & 12 & 6 & 8 & 6 & 11\tabularnewline
\hline
\makecell{Sampling Rate (Hz)} & 50 & 50 & 100 & \textbackslash{} & 200\tabularnewline
\hline
Window Size & 128 & 128 & 500 & 200 & 200\tabularnewline
\hline
Overlap & 64 & 64 & 0 & 100 & 100\tabularnewline
\hline
\end{tabular}
\end{table}

\begin{table}[t]
\renewcommand\arraystretch{1.2}
\centering

\scriptsize
\caption{Summary of key characteristics for wearable device-based HAR datasets.}\label{tab2}
\vspace{-0.3cm}

\begin{tabular}{|c|c|c|c|c|c|}
\hline
\diagbox[innerwidth=2cm]{Attributes}{Datasets} & \makecell{Oppo} & \makecell{PAMAP2} & \makecell{DSADS\\} & \makecell{RealWorld} & \makecell{SHO} \tabularnewline
\hline
Participant Count & 4 & 9 & 8 & 15 & 10\tabularnewline
\hline
Activity Number & 18 & 12 & 19 & 8 & 7 \tabularnewline
\hline
IMU Number & 7 & 3 & 5 & 7 & 5\tabularnewline
\hline
\makecell{Sampling Rate (Hz)} & 30 & 100 & 25 & 50 & 50\tabularnewline
\hline
Window Size & 48 & 48 & 125 & 48 & 48\tabularnewline
\hline
Overlap & 24 & 24 & 0 & 24 & 24\tabularnewline
\hline
\end{tabular}
\end{table}

\begin{table*}[t]
\renewcommand\arraystretch{1.2}
\centering
\footnotesize

\caption{Detailed configuration parameters for the TSF framework across different blocks.}\label{tab3}
\begin{tabular}{|c|c|c|c|c|c|c|c|c|c|}
\hline
  \multicolumn{1}{|c|}{\multirow{3}{*}{\diagbox[innerwidth=2.8cm]{Hyper-Param}{TSF Blocks}}} & \multicolumn{3}{c|}{IMU Fusion Block} & \multicolumn{3}{c|}{Modality Node Fusion Block} & \multicolumn{3}{c|}{Temporal Info Fusion Block} \\
  \cline{2-10}
  \multicolumn{1}{|c|}{} & \makecell{CConv\underline{\hspace{0.5em}}\\grav} &  \makecell{CConv\underline{\hspace{0.5em}}\\gyro} & \makecell{CConv\underline{\hspace{0.5em}}\\lacc} & \makecell{Modality \\ Projection} & \makecell{Graph\\Aggregation} & \makecell{Squeeze} & \makecell{Conv} & \makecell{Binary Mask\\Projection} & \makecell{Self-Attn}\\
\hline
  \multicolumn{1}{|c|}{Layer Number} & 1 & 1 & 1 & 1 & 2 & 128 & 1 & 1 & 2 \\
\hline
  \multicolumn{1}{|c|}{Kernel Size} & 1$\times$11 & 1$\times$10 & 1$\times$11 & 1$\times$1 & $\backslash$ & 1$\times$1 & 1$\times$5 & 1$\times$1 & $\backslash$ \\
\hline
  \multicolumn{1}{|c|}{Channel Number} & 64 & 64 & 64 & 3$\times$64$\rightarrow$96  & 96 & 3$\times$96$\rightarrow$128 & 128 & 2$\times$128$\rightarrow$2 & 128 \\
\hline
\end{tabular}
\end{table*}

\subsubsection{Evaluation Protocol}
\label{sec:5.1.3}

In the paper, evaluation metrics include the mean F1 measure (F1) and the weighted F1 (WF1), with WF1 adjusting for the proportion of classes. These metrics are standard in the field, and for further details, readers are referred to \cite{chen2021deep}.

The models are evaluated using the Leave One Subject Out Cross Validation (LOSO-CV), where data from one subject is used as the test set and the remaining data for training, rotating until all subjects are tested. This is repeated over three runs, calculating the average F1 and WF1 for each run, and the results are presented as the mean and variance of these runs.

For the SHL2018 dataset, with data from only one participant, LOSO-CV is replaced by testing on the predefined test sets. For the MobiAct dataset, with 61 participants, a 10-fold Cross-Validation (K-Fold-CV, K=10) approach is employed, randomly selecting subjects for each fold.

\subsection{Comparisons with Previous Methods}
\label{sec:5.2}

The proposed TSF framework is evaluated against a suite of nine methods: DeepConvLSTM \cite{ordonez2016deep}, DeepSense \cite{yao2017deepsense}, GlobalFusion \cite{liu2020globalfusion}, Transformer Encoder \cite{mahmud2020human}, Attend-Discriminate \cite{abedin2021attend}, DynamicWHAR \cite{miao2022towards}, ConvTransformer \cite{yang2022predictive}, IF-ConvTransformer \cite{zhang2022if}, and ConvBoost \cite{shao2023convboost}. These methods are cataloged in Table~\ref{tab4} in chronological order. DeepConvLSTM is designated as the baseline for these comparative assessments.

\textbf{(1) Recognition Performance}

\textbf{First}, TSF stands out in classification performance among its competitors, showcasing the best results in nearly all ten datasets. Notably, it surpasses the baseline method, DeepConvLSTM, improving F1/WF1 scores across these datasets by significant margins: 3.89$\%$/2.63$\%$, 3.95$\%$/3.76$\%$, 4.38$\%$/4.26$\%$, 3.91$\%$/4.05$\%$, 4.62$\%$/2.01$\%$, 11.56$\%$/2.86$\%$, 2.42$\%$/0.71$\%$, 4.93$\%$/4.78$\%$, 6.45$\%$/6.45$\%$, and 2.22$\%$/2.22$\%$, in the order presented in Table 4. These improvements are substantial and come with acceptable standard errors. Additionally, while the second-best method changes across different datasets, TSF consistently shows improvement over these methods in F1/WF1 scores, with respective increases of -0.45$\%$/0.51$\%$, 1.53$\%$/1.75$\%$, 1.44$\%$/1.04$\%$, 0.80$\%$/0.89$\%$, 1.82$\%$/0.81$\%$, 2.15$\%$/1.40$\%$, -0.06$\%$/0.06$\%$, 1.72$\%$/1.99$\%$, 0.88$\%$/0.88$\%$ and 0.10$\%$/0.10$\%$. This clearly demonstrates TSF's advantage in performance.

\textbf{Second}, the comparative performance of existing methods shows significant variation. Analysis reveals that Transformer Encoder and DynamicWHAR notably underperform the baseline method DeepConvLSTM, achieving 75.45$\%$/82.26$\%$ and 78.65$\%$/86.64$\%$ versus 83.34$\%$/90.41$\%$ in average F1/WF1 scores across ten datasets, respectively. The Transformer Encoder, being purely based on the Transformer architecture, does not adequately address sensor fusion and local context correlations. DynamicWHAR, on the other hand, treats timestamps as feature channels for compression and uses GCN to combine different sensors at the sample level, not at the timestamp level, resulting in suboptimal information fusion. Both models potentially struggle with sensor fusion and temporal association. Additionally, DeepSense and GlobalFusion, which employ STFT for data length reduction, only manage to match the performance of DeepConvLSTM, with scores of 84.04$\%$/90.45$\%$ and 83.63$\%$/90.67$\%$ versus 83.34$\%$/90.41$\%$ in averaged F1/WF1 scores on ten datasets. The STFT technique may cause significant loss of context information. In contrast, Attend-Discriminate, ConvBoost, ConvTransformer, and IF-ConvTransformer demonstrate clear superiority over DeepConvLSTM, with scores of 85.16$\%$/91.70$\%$, 86.03$\%$/92.19$\%$, 85.99$\%$/91.86$\%$, and 86.73$\%$/92.47$\%$ versus 83.34$\%$/90.41$\%$ in averaged F1/WF1 scores across ten datasets. Attend-Discriminate and ConvBoost enhance recognition performance by integrating GAT for timestamp-wise sensor fusion and data augmentation techniques, respectively. ConvTransformer replaces LSTM with Transformer for better long-term context associations, while IF-ConvTransformer further incorporates an IMU fusion block and GAT to improve sensor fusion. As a result, IF-ConvTransformer achieves the best overall performance among the compared methods.

\textbf{Finally}, TSF shows clear advantages over its predecessor, IF-ConvTransformer, with significant performance improvements (88.18$\%$/93.78$\%$ versus 86.73$\%$/92.47$\%$). This enhancement is attributed to TSF's extension of IF-ConvTransformer, incorporating adaptive filtering in the graph Fourier domain for more effective heterogeneous sensor fusion, and using adaptive wavelet frequency selection to reduce redundancy. Detailed insights into these improvements will be provided through ablation experiments in Sections~\ref{sec:5.3.2} and \ref{sec:5.3.3}. Additionally, it is noteworthy that although TSF does not achieve the top performance in all datasets (such as HAPT and PAMAP2), the difference between it and the best-performing method on these datasets is relatively small, indicating its robust performance across a range of scenarios.

\begin{table*}[t]
	\centering
	\renewcommand\arraystretch{1.4}
    \scriptsize
	\caption{Comparative analysis of classification performance across different methods, with a focus on TSF and its counterparts. The performance metrics used are the macro F-measure and the weighted F-measure, noted as F1 and WF1, respectively. The results are presented for ten distinct datasets. Each method's scores are listed along with their standard errors and the percentage increase or decrease in performance relative to the baseline method, DeepConvLSTM. The highest scores are in \textbf{boldface}, while the second-best scores are \underline{underlined}, allowing for a quick visual assessment of each method's relative performance.}\label{tab4}
	\setlength{\tabcolsep}{0.55mm}{
	\begin{tabular}{cc|c|c|c|c|c|c|c|c|c}
		\midrule[0.75pt]
		\multirow{2}*{Methods} & {HAPT}  &  {MotionSense}  & {SHL2018} & {HHAR} & {MobiAct} & {Oppo} & {Pamap2} & {RealWorld} & {DSADS} & {SHO}\\
		\cmidrule(r){2-2} \cmidrule(r){3-3} \cmidrule(r){4-4} \cmidrule(r){5-5} \cmidrule(r){6-6} \cmidrule(r){7-7} \cmidrule(r){8-8} \cmidrule(r){9-9} \cmidrule(r){10-10} \cmidrule(r){11-11}
		&F1/WF1($\%$)     &F1/WF1($\%$)     &F1/WF1($\%$)     &F1/WF1($\%$)     &F1/WF1($\%$)     &F1/WF1($\%$)     &F1/WF1($\%$)     &F1/WF1($\%$)     &F1/WF1($\%$)       &F1/WF1($\%$) \\\midrule[0.75pt]
			\multirow{3}{*}{\textbf{\shortstack{Deep \\ ConvLSTM \\ \cite{ordonez2016deep}}}}
			&82.00/92.11          &93.39/94.06           &76.93/79.12           &91.90/91.97          &87.85/94.95          &40.88/86.49          &89.91/94.20          &90.18/90.82         &83.25                &97.13                \tabularnewline
            &$\pm$1.05/$\pm$0.36  &$\pm$0.17/$\pm$0.27   &$\pm$0.31/$\pm$0.33   &$\pm$0.28/$\pm$0.30  &$\pm$0.34/$\pm$0.11  &$\pm$0.67/$\pm$0.25  &$\pm$0.08/$\pm$0.10  &$\pm$0.38/$\pm$0.31 &$\pm$0.49            &$\pm$0.75             \tabularnewline
            &(0)/(0)              &(0)/(0)               &(0)/(0)               &(0)/(0)              &(0)/(0)              &(0)/(0)              &(0)/(0)              &(0)/(0)             &(0)                  &(0)                  \tabularnewline
			\hline
            \multirow{3}{*}{\textbf{\shortstack{DeepSense \\ \cite{yao2017deepsense}}}}
			&\textbf{86.34}/93.36 &93.21/94.24           &76.02/78.51            &92.02/92.03          &89.39/95.30         &42.67/86.98	      &89.62/92.51	        &91.56/91.99	     &83.24      	       &96.34                 \tabularnewline
            &$\pm$0.74/$\pm$0.37  &$\pm$0.44/$\pm$0.39   &$\pm$0.50/$\pm$0.54    &$\pm$0.35/$\pm$0.39  &$\pm$0.13/$\pm$0.11 &$\pm$0.11/0.05	      &$\pm$2.02/$\pm$0.55	&$\pm$0.17/$\pm$0.16 &$\pm$1.11	           &$\pm$0.86             \tabularnewline
            &(4.34)/(1.25)        &(-0.18)/(0.18)        &(-0.91)/(-0.61)        &(0.12)/(0.06)        &(1.54)/(0.35)       &(1.79)/(0.49)        &(-0.29)/(-1.69)      &(1.38)/(1.17)       &(-0.01)              &(-0.79)       \tabularnewline
			\hline
            \multirow{3}{*}{\textbf{\shortstack{Global \\ Fusion \\ \cite{liu2020globalfusion}}}}
			&84.66/94.04	      &94.84/94.94	         &76.98/79.62	         &91.74/91.56	       &89.23/95.64         &37.31/85.41          &88.17/91.49          &90.82/91.41         &85.57                &97.00                \tabularnewline
    		&$\pm$0.22/$\pm$0.35  &$\pm$0.10/$\pm$0.21   &$\pm$0.30/$\pm$0.23    &$\pm$0.33/$\pm$0.34  &$\pm$0.52/$\pm$0.20 &$\pm$0.96/$\pm$0.20  &$\pm$1.28/$\pm$0.61  &$\pm$0.32/$\pm$0.16 &$\pm$0.26            &$\pm$0.48             \tabularnewline
            &(2.66)/(1.93)        &(1.45)/(0.88)         &(0.05)/(0.50)          &(-0.16)/(-0.41)      &(1.38)/(0.69)       &(-3.56)/(-1.08)      &(-1.75)/(-2.71)        &(0.63)/(0.59)     &(2.32)               &(-0.13)                \tabularnewline
			\hline
            \multirow{3}{*}{\textbf{\shortstack{Transformer \\ Encoder \\ \cite{mahmud2020human}}}}
			&81.82/91.93          &85.71/85.31           &66.77/67.38            &85.59/85.57          &80.37/90.72         &48.83/87.37          &67.99/74.39          &74.63/77.19         &80.77                &81.97                \tabularnewline
            &$\pm$0.97/$\pm$0.28  &$\pm$0.82/$\pm$0.80   &$\pm$1.11/$\pm$1.79    &$\pm$0.64/$\pm$0.58  &$\pm$0.88/$\pm$0.27 &$\pm$0.99/$\pm$0.14  &$\pm$0.96/$\pm$0.65  &$\pm$1.39/$\pm$1.26 &$\pm$1.84            &$\pm$2.35             \tabularnewline
            &(-0.18)/(-0.18)      &(-7.68)/(-8.75)       &(-10.16)/(-11.74)      &(-6.31)/(-6.40)      &(-7.48)/(-4.23)     &(7.95)/(0.88)        &(-21.92)/(-19.81)    &(-15.55)/(-13.63)   &(-2.48)              &(-15.16)              \tabularnewline
			\hline
            \multirow{3}{*}{\textbf{\shortstack{Attend- \\ Discriminate \\ \cite{abedin2021attend}}}}
			&81.95/93.40          &95.79/95.92           &78.69/80.85            &94.17/94.23          &86.43/94.88         &46.29/87.04          &92.31/94.49          &92.09/92.30         &86.12                &97.79                \tabularnewline
            &$\pm$0.32/$\pm$0.49  &$\pm$0.31/$\pm$0.10   &$\pm$0.06/$\pm$0.17    &$\pm$0.45/$\pm$0.46  &$\pm$0.23/$\pm$0.19 &$\pm$1.09/$\pm$0.14  &$\pm$1.04/$\pm$0.29  &$\pm$0.60/$\pm$0.68 &$\pm$0.64            &$\pm$0.40             \tabularnewline
            &(-0.05)/(1.29)       &(2.40)/(1.86)         &(1.76)/(1.73)          &(2.27)/(2.26)        &(-1.42)/(-0.07)     &(5.41)/(0.55)        &(2.40)/(0.29)        &(1.91)/(1.48)       &(2.87)               &(0.66)                \tabularnewline
			\hline
            \multirow{3}{*}{\textbf{\shortstack{Dynamic- \\ WHAR \\ \cite{miao2022towards}}}}
			&83.12/93.58          &91.12/90.39           &68.25/69.48            &89.06/89.22          &84.26/93.50         &28.41/82.39          &79.69/84.93          &92.62/93.03         &73.00	               &96.97                \tabularnewline
            &$\pm$0.36/$\pm$0.1	  &$\pm$0.31/$\pm$0.32   &$\pm$0.45/$\pm$0.22    &$\pm$0.17/$\pm$0.19  &$\pm$0.10/$\pm$0.17 &$\pm$0.61/$\pm$0.23  &$\pm$0.35/0.66       &$\pm$0.26/$\pm$0.22 &$\pm$0.42            &$\pm$0.64            \tabularnewline
            &(1.12)/(1.47)        &(-2.27)/(-3.67)       &(-8.68)/(-9.64)        &(-2.84)/(-2.75)      &(-3.59)/(-1.45)     &(-12.47)/(-4.1)      &(-10.22)/(-9.27)     &(2.44)/(2.21)       &(-10.25)             &(-0.16)      \tabularnewline
			\hline
            \multirow{3}{*}{\textbf{\shortstack{Conv- \\ Transformer \\ \cite{yang2022predictive}}}}
			&83.97/93.41          &95.62/96.01           &\underline{79.87}/\underline{82.34}            &93.15/93.11          &\underline{90.65}/96.02         &49.47/87.84          &\textbf{92.39}/\underline{94.85}          &91.44/91.61         &85.5                 &97.9                 \tabularnewline
            &$\pm$0.36/$\pm$0.1	  &$\pm$0.31/$\pm$0.32   &$\pm$0.45/$\pm$0.22    &$\pm$0.17/$\pm$0.19  &$\pm$0.10/$\pm$0.17 &$\pm$0.61/$\pm$0.23  &$\pm$0.35/0.66       &$\pm$0.26/$\pm$0.22 &$\pm$0.42            &$\pm$0.64            \tabularnewline
            &(1.97)/(1.3)	      &(2.23)/(1.95)	     &(2.94)/(3.22)	         &(1.25)/(1.14)	       &(2.8)/(1.07)	    &(8.59)/(1.35)	      &(2.48)/(0.65)	    &(1.26)/(0.79)	     &(2.25)       	       &(0.77)               \tabularnewline
			\hline
            \multirow{3}{*}{\textbf{\shortstack{IF-Conv- \\ Transformer \\ \cite{zhang2022if}}}}
			&85.35/\underline{94.23}          &\underline{95.81}/\underline{96.07}           &78.67/81.36            &\underline{95.01}/\underline{95.13}          &90.01/95.66         &\underline{50.29}/\underline{87.95}          &92.21/93.86          &92.81/93.29         &\underline{88.82}                &98.28                \tabularnewline
            &$\pm$1.03/$\pm$0.43  &$\pm$0.15/$\pm$0.17   &$\pm$0.65/$\pm$0.77    &$\pm$0.52/$\pm$0.51  &$\pm$0.55/$\pm$0.11 &$\pm$2.09/$\pm$0.54  &$\pm$0.73/$\pm$0.39  &$\pm$0.11/$\pm$0.13 &$\pm$0.41            &$\pm$0.46             \tabularnewline
            &(3.35)/(2.12)        &(2.42)/(2.01)         &(1.74)/(2.24)          &(3.11)/(3.16)        &(2.16)/(0.71)       &(9.41)/(1.46)        &(2.30)/(-0.34)       &(2.63)/(2.47)       &(5.57)               &(1.15)                \tabularnewline
			\hline
            \multirow{3}{*}{\textbf{\shortstack{ConvBoost \\ \cite{shao2023convboost}}}}
			&85.17/93.59          &94.04/93.93           &79.47/81.90             &94.04/93.93          &89.93/\underline{96.15}         &45.94/87.76          &91.83/94.54          &\underline{93.39}/\underline{93.61}         &87.26                &\underline{99.25}           \tabularnewline
            &$\pm$0.17/$\pm$0.44  &$\pm$0.52/$\pm$0.56	 &$\pm$0.48/$\pm$0.50    &$\pm$0.52/$\pm$0.56  &$\pm$0.48/$\pm$0.21	&$\pm$1.54/$\pm$0.21  &$\pm$1.60/$\pm$0.28	&$\pm$0.23/$\pm$0.29 &$\pm$0.46            &$\pm$0.06             \tabularnewline
            &(3.17)/(1.48)	      &(0.65)/(-0.13)	     &(2.54)/(2.78)	         &(2.14)/(1.96)	       &(2.08)/(1.2)	    &(5.06)/(1.27)	      &(1.92)/(0.34)	    &(3.21)/(2.79)	     &(4.01)	           &(2.12)
                \tabularnewline
			\hline
			\hline
            \multirow{3}{*}{\textbf{\shortstack{TSF}}}
			&\underline{85.89}/\textbf{94.74}          &\textbf{97.34}/\textbf{97.82}           &\textbf{81.31}/\textbf{83.38}            &\textbf{95.81}/\textbf{96.02}          &\textbf{92.47}/\textbf{96.96}         &\textbf{52.44}/\textbf{89.35}          &\underline{92.33}/\textbf{94.91}          &\textbf{95.11}/\textbf{95.60}         &\textbf{89.70}                &\textbf{99.35}                \tabularnewline
            &$\pm$0.56/$\pm$0.21  &$\pm$0.21/$\pm$0.18   &$\pm$0.16/$\pm$0.30    &$\pm$0.31/$\pm$0.24  &$\pm$0.37/$\pm$0.12 &$\pm$1.04/$\pm$0.31  &$\pm$0.81/$\pm$0.29  &$\pm$0.11/$\pm$0.10 &$\pm$1.15            &$\pm$0.07             \tabularnewline
            &(3.89)/(2.63)        &(3.95)/(3.76)         &(4.38)/(4.26)          &(3.91)/(4.05)        &(4.62)/(2.01)       &(11.56)/(2.86)       &(2.42)/(0.71)        &(4.93)/(4.78)       &(6.45)               &(2.22)                \tabularnewline
			\midrule[0.75pt]
	\end{tabular}}

\end{table*}

\textbf{(2) Computational Complexity and Parameter Count}

\begin{figure}[!htbp]
\centering
\footnotesize
    \begin{minipage}[b]{0.5\textwidth}
        \centering
        \subfloat[The FLOPs of different models across multiple datasets]{
        \includegraphics[width=7.6cm]{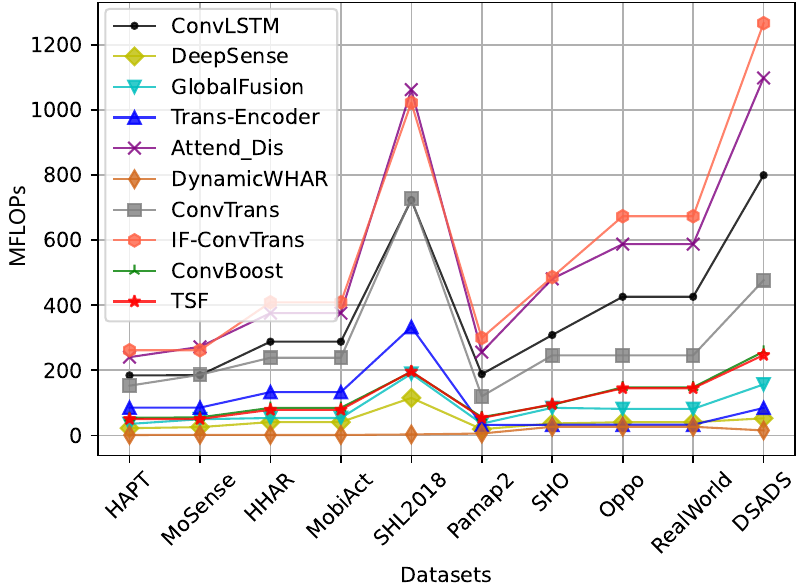}
        \label{fig5_a}
        }
    \vspace{0.2cm}
    \end{minipage}
    \begin{minipage}[b]{0.5\textwidth}
        \centering
        \subfloat[The number of network parameters of different models across multiple datasets]{
        \includegraphics[width=7.6cm]{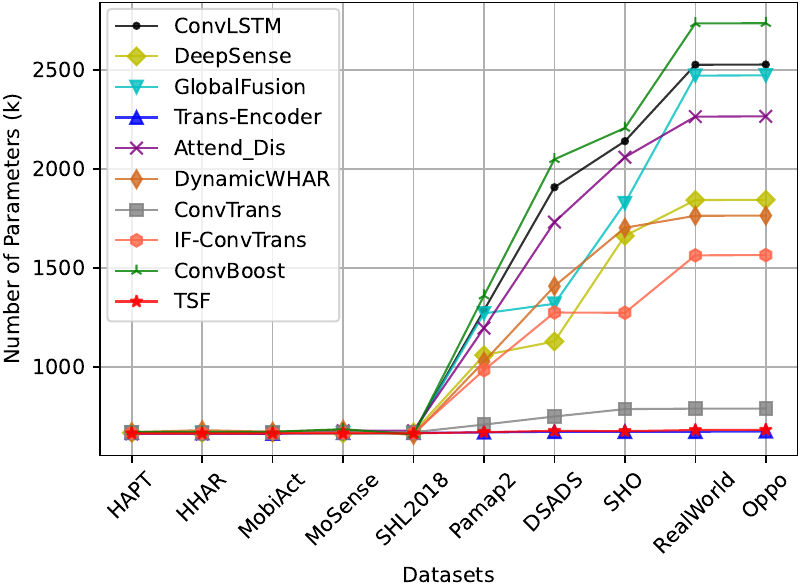}
        \label{fig5_b}
        }
    \end{minipage}
\caption{Comparative analysis of computational complexity and model size across different methods and datasets: (a) Graph illustrating the FLOPs for various deep models across multiple datasets. (b) Graph showing the variation in the number of parameters for these models, indicating model size and potential complexity.}
\label{fig5}
\end{figure}

Figure~\ref{fig5_a} illustrates the Floating Point Operations Per Second (FLOPs) of TSF and various existing methods across different datasets. In general, TSF exhibits higher FLOPs when compared to Transformer Encoder (in the case of wearable device-based datasets) and DynamicWHAR. Transformer Encoder is a purely Transformer-based model with limited sensor fusion designs, while DynamicWHAR directly compresses timestamp channels for context correlation. Although these two models employ highly efficient network architectures, they suffer from significant performance degradation due to their limited capabilities in sensor fusion and context association, as evident in Table~\ref{tab4}.

Additionally, while the FLOPs of TSF are generally comparable with those of DeepSense and GlobalFusion, they are still higher on some datasets such as SHL2018, Oppo, and DSADS. These two models achieve a drastic reduction in the length of input data through STFT preprocessing. However, as discussed in Section~\ref{sec:ti-fusion}, this computational advantage comes at the expense of losing context information, resulting in unsatisfactory recognition performance.

It is important to note that TSF, through its adaptive frequency selection mechanism, effectively mitigates context redundancies, thereby significantly reducing the FLOPs when compared to its previous version, IF-ConvTransformer. In comparison to other models, TSF strikes an effective balance between computational complexity and recognition performance.

Figure~\ref{fig5_b} provides insight into the parameter count of TSF, which remains relatively stable even as the number of sensors increases. In contrast, most existing methods, such as ConvLSTM and ConvBoost, exhibit a sharp rise in the number of parameters as the sensor count grows. This phenomenon is attributed to the adaptive graph Fourier filtering used in TSF, eliminating the need for commonly employed feature compression layers \cite{ordonez2016deep,yao2017deepsense,liu2020globalfusion,abedin2021attend,miao2022towards,zhang2022if} for extensive multi-sensor fusion. As a result, TSF achieves a significant reduction in the overall number of parameters.

In summary, TSF effectively strikes a balance between its recognition performance, computational complexity, and the number of model parameters.

\subsection{Ablation Study}
\label{sec:5.3}

This section presents a series of ablation studies on the TSF framework. Specifically, it explores the effectiveness of the IMU fusion block, the modality node fusion block, and the temporal information fusion block, with their respective evaluations detailed in Sections~\ref{sec:5.3.1}, \ref{sec:5.3.2}, and \ref{sec:5.3.3}.

\subsubsection{IMU Fusion Block Evaluation}
\label{sec:5.3.1}

This subsection details a comparative analysis of the IMU fusion block within the TSF framework against its various variants. These modifications involve selectively omitting key components of the IMU fusion block while maintaining the integrity of the remaining elements of the TSF. The results of these ablation studies are tabulated in Table~\ref{tab5}.

\textbf{(1) Effectiveness of IMU Fusion Blocks}

The performance of the IMU fusion blocks is assessed by substituting them with one-dimensional convolutional layers, which yields the TSF-IMU (noIMUFusion) configuration, as shown in Table~\ref{tab5}. It is evident that the TSF-IMU (noIMUFusion) model experiences an average performance drop of 2.34$\%$/1.67$\%$ in F1 and WF1 scores, respectively, across ten datasets when compared to the original TSF model  (85.84$\%$/92.11$\%$ vs. 88.18$\%$/93.78$\%$). This decline in performance highlights the significant role of the IMU fusion block in the framework.

\textbf{(2) Adaptive Complementary Filter versus Traditional Complementary Filter}

An alternative TSF-IMU model is created by substituting the adaptive complementary filter-based posture branch with a traditional complementary filter, named TSF-IMU (Complementary Filter). This model exhibits a notable average reduction of 3.37$\%$/2.34$\%$ in F1 and WF1 scores, respectively, across ten datasets (84.81$\%$/91.44$\%$ vs. 88.18$\%$/93.78$\%$), when contrasted with the standard TSF. Furthermore, the TSF-IMU (Complementary Filter) performs slightly worse than the TSF-IMU (noIMUFusion) variant (84.81$\%$/91.44$\%$ vs. 85.84$\%$/92.11$\%$). This inferior performance is attributed to the traditional complementary filter's reliance on static parameters for sensor fusion, which lacks the data adaptability, as discussed in Section~\ref{sec:4.2.1}. In contrast, the adaptive complementary filter can adjust to the data's unique features, leading to improved results.

\textbf{(3) Impact of Removing Posture Sensor Attention}

Posture sensor attention plays an important role in the IMU fusion block as it sets the cut-off frequency for both the high-pass and low-pass complementary filters, a detail has been explained in Section~\ref{sec:4.2.1}. Removing this posture sensor attention leads to a simplification of the IMU fusion block into a basic multi-branch subnet, forming the TSF-IMU (noSensorAttn) model. Table~\ref{tab5} illustrates that the TSF-IMU (noSensorAttn) model experiences an average performance drop of 1.54$\%$/1.26$\%$ compared to the original TSF model (86.64$\%$/92.52$\%$ vs. 88.18$\%$/93.78$\%$). Moreover, a decline in performance is observed across all datasets used in the study for the TSF-IMU (noSensorAttn) model. These experimental results demonstrate the vital importance of including posture sensor attention in the model.

\textbf{(4) Impact of Varying Kernel Sizes}

In the context of adaptive complementary filtering, large kernels function as sliding cut-off windows, eliminating timestamps that are far apart (as detailed in Section~\ref{sec:4.2.2}). This section of the paper compares models with various kernel sizes (namely 5, 21, 31) against the TSF model, which uses a kernel size of 11.  According to the results in Table~\ref{tab5}, these models exhibit a respective decrease of 0.37$\%$/0.34$\%$, 0.85$\%$/0.45$\%$, and 1.30$\%$/0.58$\%$ in average F1/WF1 scores across ten datasets (with scores of 87.81$\%$/93.44$\%$, 87.33$\%$/93.33$\%$, and 86.88$\%$/93.20$\%$ compared to TSF's 88.18$\%$/93.78$\%$). This variation in performance can be attributed to the size of the kernels: too small a kernel might not cover essential historical timestamps adequately, while too large a kernel could cause a `drift' issue during the integration process.

\begin{table*}[t]
	\centering
	\renewcommand\arraystretch{1.4}
    \scriptsize
	\caption{Detailed ablation study results highlighting the impact of various modifications to the IMU fusion block on F1 and WF1 scores across multiple datasets.}\label{tab5}
	\setlength{\tabcolsep}{0.68mm}{
	\begin{tabular}{cc|c|c|c|c|c|c|c|c|c}
		\midrule[0.75pt]
		\multirow{2}*{Methods} & {HAPT}  &  {MotionSense}  & {SHL2018} & {HHAR} & {MobiAct} & {Oppo} & {Pamap2} & {RealWorld} & {DSADS} & {SHO}\\
		\cmidrule(r){2-2} \cmidrule(r){3-3} \cmidrule(r){4-4} \cmidrule(r){5-5} \cmidrule(r){6-6} \cmidrule(r){7-7} \cmidrule(r){8-8} \cmidrule(r){9-9} \cmidrule(r){10-10} \cmidrule(r){11-11}
		&F1/WF1($\%$)     &F1/WF1($\%$)     &F1/WF1($\%$)     &F1/WF1($\%$)     &F1/WF1($\%$)     &F1/WF1($\%$)     &F1/WF1($\%$)     &F1/WF1($\%$)     &F1/WF1($\%$)       &F1/WF1($\%$) \\\midrule[0.75pt]
			\multirow{3}{*}{\textbf{\shortstack{TSF-IMU \\ (noIMU \\ Fusion)}}}
			&84.75/94.52          &97.10/97.37           &80.73/82.72           &95.10/95.12          &91.81/96.63          &43.66/86.67          &89.43/92.23          &92.18/92.21         &84.68      &98.91                \tabularnewline
            &$\pm$0.82/$\pm$0.12  &$\pm$0.11/$\pm$0.06   &$\pm$0.40/$\pm$0.33   &$\pm$0.16/$\pm$0.14  &$\pm$0.43/$\pm$0.26  &$\pm$0.20/$\pm$0.17  &$\pm$2.21/$\pm$0.45  &$\pm$0.18/$\pm$0.13 &$\pm$0.40  &$\pm$0.11            \tabularnewline
            &(0)/(0)              &(0)/(0)               &(0)/(0)               &(0)/(0)              &(0)/(0)              &(0)/(0)              &(0)/(0)              &(0)/(0)             &(0)        &(0)                  \tabularnewline
			\hline
            \multirow{3}{*}{\textbf{\shortstack{TSF-IMU \\ (Complementary \\ Filter)}}}
			&84.67/\textbf{94.85}          &96.62/97.04           &81.14/83.22           &94.71/95.02          &91.31/96.60          &44.43/87.09          &87.97/92.44          &89.95/90.81         &85.58      &91.73                \tabularnewline
            &$\pm$0.74/$\pm$0.37  &$\pm$0.24/$\pm$0.13   &$\pm$0.10/$\pm$0.11   &$\pm$0.45/$\pm$0.43  &$\pm$0.28/$\pm$0.03  &$\pm$0.55/$\pm$0.06  &$\pm$1.40/$\pm$0.11  &$\pm$0.20/$\pm$0.26 &$\pm$1.39  &$\pm$0.09            \tabularnewline
            &(-0.08)/(0.33)       &(-0.48)/(-0.33)       &(0.41)/(0.50)         &(-0.39)/(-0.10)      &(-0.50)/(-0.03)      &(0.77)/(0.42)        &(-1.46)/(-0.21)      &(-2.23)/(-1.40)     &(0.90)     &(-7.18)              \tabularnewline
			\hline
            \multirow{3}{*}{\textbf{\shortstack{TSF-IMU \\ (noSensorAttn)}}}
			&84.70/94.68          &97.06/97.45           &80.23/81.73           &95.28/95.35          &91.91/94.64          &46.82/87.70          &91.00/93.88          &94.02/94.36         &86.67      &98.75                \tabularnewline
    		&$\pm$0.63/$\pm$0.11  &$\pm$0.14/$\pm$0.04   &$\pm$0.32/$\pm$0.53   &$\pm$0.25/$\pm$0.26  &$\pm$0.09/$\pm$0.28  &$\pm$1.79/$\pm$0.18  &$\pm$0.90/$\pm$0.19  &$\pm$0.08/$\pm$0.12 &$\pm$0.28  &$\pm$0.36            \tabularnewline
            &(-0.05)/(0.16)       &(-0.04)/(0.08)        &(-0.50)/(-0.99)       &(0.18)/(0.23)        &(0.10)/(0.01)        &(3.16)/(1.03)        &(1.57)/(1.65)        &(1.84)/(2.15)       &(1.99)     &(-0.16)              \tabularnewline
			\hline
            \multirow{3}{*}{\textbf{\shortstack{TSF-IMU \\ (Kernel-5)}}}
			&85.37/94.75          &96.79/97.32           &80.48/82.23           &95.55/95.79          &92.08/\underline{96.93}          &\textbf{52.74}/\underline{89.19}          &\underline{91.91}/\underline{94.58}          &\underline{94.74}/\underline{95.15}         &\underline{89.04}      &\textbf{99.42}                \tabularnewline
            &$\pm$0.13/$\pm$0.32  &$\pm$0.19/$\pm$0.10   &$\pm$0.33/$\pm$0.21   &$\pm$0.26/$\pm$0.28  &$\pm$0.12/$\pm$0.13  &$\pm$0.51/$\pm$0.26  &$\pm$0.68/$\pm$0.25  &$\pm$0.38/$\pm$0.39 &$\pm$0.12  &$\pm$0.09            \tabularnewline
            &(0.62)/(0.23)        &(-0.31)/(-0.05)       &(-0.25)/(-0.49)       &(0.45)/(0.67)        &(0.27)/(0.30)        &(9.08)/(2.52)        &(2.48)/(2.35)        &(2.56)/(2.94)       &(4.36)    &(0.51)               \tabularnewline
			\hline
            \multirow{3}{*}{\textbf{\shortstack{TSF-IMU \\ (Kernel-21)}}}
			&85.28/94.71          &97.28/97.63           &\textbf{81.55}/\textbf{83.48}           &95.62/95.86          &92.00/96.77          &49.88/88.68          &89.75/93.77          &94.58/95.02         &88.13      &99.23                \tabularnewline
            &$\pm$0.32/$\pm$0.49  &$\pm$0.31/$\pm$0.10   &$\pm$0.54/$\pm$0.59   &$\pm$0.04/$\pm$0.09  &$\pm$0.31/$\pm$0.15  &$\pm$1.92/$\pm$0.32  &$\pm$0.67/$\pm$0.10  &$\pm$0.28/$\pm$0.24 &$\pm$0.62  &$\pm$0.34            \tabularnewline
            &(0.53)/(0.19)        &(0.18)/(0.26)         &(0.82)/(0.76)         &(0.52)/(0.74)        &(0.19)/(0.14)        &(6.21)/(2.01)        &(0.32)/(1.54)        &(2.40)/(2.81)       &(3.45)     &(0.32)               \tabularnewline
			\hline
            \multirow{3}{*}{\textbf{\shortstack{TSF-IMU \\ (Kernel-31)}}}
			&\underline{85.58}/94.58          &\textbf{97.55}/\textbf{97.90}           &\underline{81.33}/83.18           &\underline{95.71}/\underline{95.96}          &\underline{92.41}/96.83          &48.19/88.64          &87.16/93.52          &94.61/95.09         &87.17      &99.12                \tabularnewline
            &$\pm$0.51/$\pm$0.26  &$\pm$0.27/$\pm$0.24   &$\pm$0.54/$\pm$0.50   &$\pm$0.03/$\pm$0.02  &$\pm$0.09/$\pm$0.10  &$\pm$2.48/$\pm$0.56  &$\pm$0.18/$\pm$0.17  &$\pm$0.13/$\pm$0.06 &$\pm$0.43  &$\pm$0.10            \tabularnewline
            &(0.83)/(0.06)        &(0.45)/(0.53)         &(0.60)/(0.46)         &(0.61)/(0.84)        &(0.60)/(0.20)        &(4.53)/(1.97)        &(-2.27)/(1.29)        &(2.43)/(2.88)       &(2.49)     &(0.21)               \tabularnewline
			\hline
			\hline
            \multirow{3}{*}{\textbf{\shortstack{TSF}}}
			&\textbf{85.89}/\underline{94.74}          &\underline{97.34}/\underline{97.82}           &81.31/\underline{83.38}           &\textbf{95.81}/\textbf{96.02}          &\textbf{92.47}/\textbf{96.96}          &\underline{52.44}/\textbf{89.35}          &\textbf{92.33}/\textbf{94.91}          &\textbf{95.11}/\textbf{95.60}         &\textbf{89.70}      &\underline{99.35}                \tabularnewline
            &$\pm$0.56/$\pm$0.21  &$\pm$0.21/$\pm$0.18   &$\pm$0.16/$\pm$0.30    &$\pm$0.31/$\pm$0.24  &$\pm$0.37/$\pm$0.12 &$\pm$1.04/$\pm$0.31  &$\pm$0.81/$\pm$0.29  &$\pm$0.11/$\pm$0.10 &$\pm$1.15            &$\pm$0.07             \tabularnewline
            &(1.14)/(0.22)        &(0.24)/(0.45)         &(0.58)/(0.65)         &(0.71)/(0.90)        &(0.66)/(0.33)        &(8.78)/(2.68)        &(2.90)/(2.68)        &(2.93)/(3.39)       &(5.02)     &(0.44)               \tabularnewline
			\midrule[0.75pt]
	\end{tabular}}

\end{table*}

\subsubsection{Modality Node Fusion Block Evaluation}
\label{sec:5.3.2}

This subsection focuses on ablation experiments conducted on the modality node fusion block within the TSF framework. These experiments involved progressively removing the entire block, different filters, and the dynamic adjacency matrix from TSF, while keeping the remaining components unchanged. The results of these ablation studies are shown in Table~\ref{tab6}.

\textbf{(1) Effectiveness of the Modality Node Fusion Block}

The impact of the modality node fusion block is assessed by substituting this block with an MLP layer, resulting in the TSF (noGraphFusion) model. In this model, the MLP layer fuses nodes by directly compressing the channels of the concatenated node features. When compared with the standard TSF, the TSF (noGraphFusion) model shows an average decline of 1.20$\%$/0.64$\%$ in F1 and WF1 scores across ten datasets (86.98$\%$/93.14$\%$ vs. 88.18$\%$/93.78$\%$). This performance drop is attributed to the MLP's characteristics as an all-pass filter \cite{wang2022survey}, where it struggles to balance the proportion of homogeneous and heterogeneous information.

\textbf{(2) Role of Low-Pass and High-pass Filters}

The high-pass filters in the TSF framework are essential for capturing the heterogeneous information of nodes by preserving their unique characteristics. To evaluate the effectiveness of these high-pass filters, they were removed from the TSF model, creating the TSF-Graph (Low-pass Filter) model. This alteration essentially reduces the adaptive graph Fourier filtering to a GAT subnet, primarily focusing on homogeneous node information. As indicated in Table~\ref{tab6}, the TSF-Graph (Low-pass Filter) model shows an average decline of 0.65$\%$/0.53$\%$ in F1 and WF1 scores compared to the standard TSF (87.53$\%$/93.25$\%$ vs. 88.18$\%$/93.78$\%$). Similarly, removing the low-pass filters from TSF leads to the TSF-Graph (High-pass Filter) model. This model also shows an average decrease of 0.56$\%$/0.49$\%$ in F1 and WF1 scores (87.62$\%$/93.29$\%$ vs. 88.18$\%$/93.78$\%$) when compared with the original TSF. These results highlight the significance of both low-pass and high-pass filters in the TSF framework.

Additionally, it is noted that the TSF-Graph (High-pass Filter) performs marginally better than the TSF-Graph (Low-pass Filter) in the context of five smartphone-based datasets (90.18$\%$/93.41$\%$ vs. 89.99$\%$/93.24$\%$). This is attributed to the nature of these datasets, recorded using a single IMU and thus representing a graph of a posture node and a motion node, which are largely heterogeneous due to their distinct physical meanings. Conversely, for the wearable device-based datasets, where data is recorded from IMUs placed at multiple body positions and strong correlations between different body parts are typical, the TSF-Graph (High-pass Filter) shows slightly inferior performance compared to the TSF-Graph (Low-pass Filter) (85.05$\%$/93.15$\%$ v.s. 85.08$\%$/93.26$\%$). These variations in performance aligns with the expected behavior in HAR tasks, reflecting the dominance of heterogeneous and homogeneous node information in different scenarios.

\textbf{(3) Effect of Static versus Dynamic Adjacency Graphs}

The dynamic adjacency graph's impact is assessed by setting the adjacency matrix $\bm{A}$, as defined in Equation~\ref{e4.11}, to a static matrix entirely filled with ones. The results, shown in Table~\ref{tab6}, indicate that the TSF-Graph model without the dynamic graph component (TSF-Graph (noDynamicGraph)) experiences an average performance reduction of 0.86$\%$/0.61$\%$ in F1/WF1 scores compared to the TSF (87.32$\%$/93.17$\%$ vs. 88.18$\%$/93.78$\%$). This decrease in performance can be attributed to the fact that node correlations change according to the data characteristics and different timestamps, while static adjacency cannot adjust to these changes effectively. On the other hand, a dynamic adjacency graph is designed to adjust edge weights adaptively, leading to superior performance.

\begin{table*}[t]
	\centering
	\renewcommand\arraystretch{1.4}
    \scriptsize
	\caption{Results from ablation studies on the modality node fusion block, comparing F1 and WF1 scores across various datasets for different configurations of the TSF-Graph model.}\label{tab6}
	\setlength{\tabcolsep}{0.68mm}{
	\begin{tabular}{cc|c|c|c|c|c|c|c|c|c}
		\midrule[0.75pt]
		\multirow{2}*{Methods} & {HAPT}  &  {MotionSense}  & {SHL2018} & {HHAR} & {MobiAct} & {Oppo} & {Pamap2} & {RealWorld} & {DSADS} & {SHO}\\
		\cmidrule(r){2-2} \cmidrule(r){3-3} \cmidrule(r){4-4} \cmidrule(r){5-5} \cmidrule(r){6-6} \cmidrule(r){7-7} \cmidrule(r){8-8} \cmidrule(r){9-9} \cmidrule(r){10-10} \cmidrule(r){11-11}
		&F1/WF1($\%$)     &F1/WF1($\%$)     &F1/WF1($\%$)     &F1/WF1($\%$)     &F1/WF1($\%$)     &F1/WF1($\%$)     &F1/WF1($\%$)     &F1/WF1($\%$)     &F1/WF1($\%$)       &F1/WF1($\%$) \\\midrule[0.75pt]
			\multirow{3}{*}{\textbf{\shortstack{TSF-Graph \\ (noGraph \\ Fusion)}}}
			&\underline{84.87}/94.31          &\underline{97.25}/\underline{97.68}           &\textbf{81.61}/\underline{82.67}           &\textbf{95.86}/\underline{96.01}          &92.06/\underline{96.89}          &47.03/87.96           &89.36/93.60          &94.09/94.53          &\underline{88.85}      &98.83                \tabularnewline
            &$\pm$0.09/$\pm$0.30  &$\pm$0.03/$\pm$0.04   &$\pm$0.39/$\pm$0.28   &$\pm$0.14/$\pm$0.11  &$\pm$0.37/$\pm$0.15  &$\pm$1.12/$\pm$0.33   &$\pm$1.59/$\pm$0.33  &$\pm$0.12/$\pm$0.18  &$\pm$0.46  &$\pm$0.47            \tabularnewline
            &(0)/(0)              &(0)/(0)               &(0)/(0)               &(0)/(0)              &(0)/(0)              &(0)/(0)               &(0)/(0)              &(0)/(0)              &(0)        &(0)                  \tabularnewline
			\hline
            \multirow{3}{*}{\textbf{\shortstack{TSF-Graph \\ (Low-pass \\ Filter)}}}
			&84.52/94.23          &97.05/97.54           &80.97/82.03           &95.36/95.58          &92.04/96.83          &50.88/88.85           &\underline{91.97}/\underline{94.48}          &94.57/94.99          &88.77      &\underline{99.19}                \tabularnewline
    		&$\pm$0.35/$\pm$0.16  &$\pm$0.05/$\pm$0.12   &$\pm$0.12/$\pm$0.08   &$\pm$0.08/$\pm$0.15  &$\pm$0.11/$\pm$0.21  &$\pm$0.15/$\pm$0.04   &$\pm$0.75/$\pm$0.07  &$\pm$0.03/$\pm$0.04  &$\pm$0.45  &$\pm$0.15            \tabularnewline
            &(-0.35)/(-0.08)        &(-0.20)/(-0.14)        &(-0.64)/(-0.64)        &(-0.50)/(-0.05)      &(-0.02)/(-0.06)        &(3.85)/(0.89)         &(2.61)/(0.88)       &(0.48)/(0.46)      &(-0.08)     &(0.36)               \tabularnewline
			\hline
            \multirow{3}{*}{\textbf{\shortstack{TSF-Graph \\ (High-pass \\ Filter)}}}
			&84.84/94.52          &96.93/97.46           &81.47/82.60           &95.41/95.62          &\underline{92.25}/96.84          &51.16/88.91           &91.59/93.96          &\underline{94.91}/\underline{95.32}          &88.52      &99.08                \tabularnewline
            &$\pm$0.57/$\pm$0.29  &$\pm$0.14/$\pm$0.13   &$\pm$0.22/$\pm$0.19   &$\pm$0.19/$\pm$0.18  &$\pm$0.11/$\pm$0.08  &$\pm$1.31/$\pm$0.23   &$\pm$0.99/$\pm$0.22  &$\pm$0.06/$\pm$0.05  &$\pm$1.65  &$\pm$0.25            \tabularnewline
            &(-0.03)/(0.21)      &(-0.32)/(-0.22)       &(-0.14)/(-0.07)         &(-0.45)/(-0.41)      &(0.19)/(-0.05)        &(4.13)/(0.95)         &(2.23)/(0.36)      &(0.82)/(0.79)       &(-0.33)    &(0.25)               \tabularnewline
			\hline
            \multirow{3}{*}{\textbf{\shortstack{TSF-Graph \\ (noDynamic \\ Graph)}}}
			&84.63/\underline{94.63}          &96.96/97.42           &80.33/82.23           &95.55/95.76          &91.95/96.76          &\underline{51.30}/\underline{88.99}           &90.71/93.69          &94.58/95.03          &88.21      &99.05                \tabularnewline
            &$\pm$0.62/$\pm$0.28  &$\pm$0.06/$\pm$0.04   &$\pm$0.37/$\pm$0.29   &$\pm$0.17/$\pm$0.19  &$\pm$0.04/$\pm$0.18  &$\pm$0.71/$\pm$0.12   &$\pm$1.38/$\pm$0.08  &$\pm$0.28/$\pm$0.24  &$\pm$0.38  &$\pm$0.16            \tabularnewline
            &(-0.24)/(0.32)      &(-0.29)/(-0.26)       &(-1.28)/(-0.44)       &(-0.31)/(-0.25)      &(-0.11)/(-0.13)        &(4.27)/(1.03)         &(1.35)/(0.09)      &(0.49)/(0.50)      &(-0.64)    &(0.22)               \tabularnewline
			\hline
			\hline
            \multirow{3}{*}{\textbf{\shortstack{TSF}}}
			&\textbf{85.89}/\textbf{94.74}          &\textbf{97.34}/\textbf{97.82}           &\underline{81.31}/\textbf{83.38}           &\underline{95.81}/\textbf{96.02}          &\textbf{92.47}/\textbf{96.96}          &\textbf{52.44}/\textbf{89.35}           &\textbf{92.33}/\textbf{94.91}          &\textbf{95.11}/\textbf{95.60}          &\textbf{89.70}      &\textbf{99.35}                \tabularnewline
            &$\pm$0.56/$\pm$0.21  &$\pm$0.21/$\pm$0.18   &$\pm$0.16/$\pm$0.30    &$\pm$0.31/$\pm$0.24  &$\pm$0.37/$\pm$0.12 &$\pm$1.04/$\pm$0.31  &$\pm$0.81/$\pm$0.29  &$\pm$0.11/$\pm$0.10 &$\pm$1.15            &$\pm$0.07             \tabularnewline
            &(1.02)/(0.43)        &(0.09)/(0.14)           &(-0.30)/(0.71)         &(-0.05)/(0.01)      &(0.41)/(0.07)        &(5.41)/(1.39)         &(2.97)/(1.31)        &(1.02)/(1.07)        &(0.85)     &(0.52)               \tabularnewline
			\midrule[0.75pt]
	\end{tabular}}

\end{table*}

\subsubsection{Temporal Information Fusion Block Evaluation}
\label{sec:5.3.3}

This section discusses the significance of the adaptive wavelet frequency selection mechanism within the temporal information fusion block. The ablation studies, with the findings detailed in Table~\ref{tab7}, examine the consequences of removing this mechanism and its core components from the temporal information fusion blocks.

\textbf{(1) Importance of Adaptive Wavelet Frequency Selection}

The effectiveness of the adaptive wavelet frequency selection is demonstrated by its omission from the temporal information fusion blocks, leading to the creation of the TSF-Context (noDWT) model. The temporal information fusion blocks now function as a ConvTransformer subnet without the ability to reduce feature length. In comparison to the original TSF, the TSF-Context (noDWT) exhibits a mean decline of 0.92$\%$/0.73$\%$ in F1/WF1 scores across ten datasets (87.26$\%$/93.05$\%$ vs. 88.18$\%$/93.78$\%$). Further examination of the TSF-Context (noLocalDWT) and TSF-Context (noGlobalDWT) models, which lack adaptive frequency selection in local and global fusion blocks respectively, reveals slightly improved performance over the TSF-Context (noDWT) model (87.71$\%$/93.34$\%$, 87.61$\%$/93.17$\%$ vs. 87.26$\%$/93.05$\%$), but still not reaching the performance of the original TSF model, as shown in Table~\ref{tab7}.

In essence, TSF outperforms on smartphone-based datasets when compared to its variants without adaptive wavelet frequency selection. Specifically, TSF leads to an average improvement of 1.38$\%$/0.93$\%$, 0.71$\%$/0.49$\%$, and 1.09$\%$/0.88$\%$ in F1/WF1 scores over TSF-Context (noDWT), TSF-Context (noLocalDWT), and TSF-Context (noGlobalDWT) respectively on these datasets. This outperformance is more pronounced than on wearable device-based datasets, where the gains are 0.44$\%$/0.53$\%$, 0.24$\%$/0.40$\%$, and 0.03$\%$/0.34$\%$ respectively. The larger segmentation windows in smartphone datasets, which could introduce more redundant context information, are managed more efficiently by TSF compared to the variant models.

\begin{figure}[t]
\centering
\footnotesize
\begin{tabular}{c}

\includegraphics[width=7.6cm]{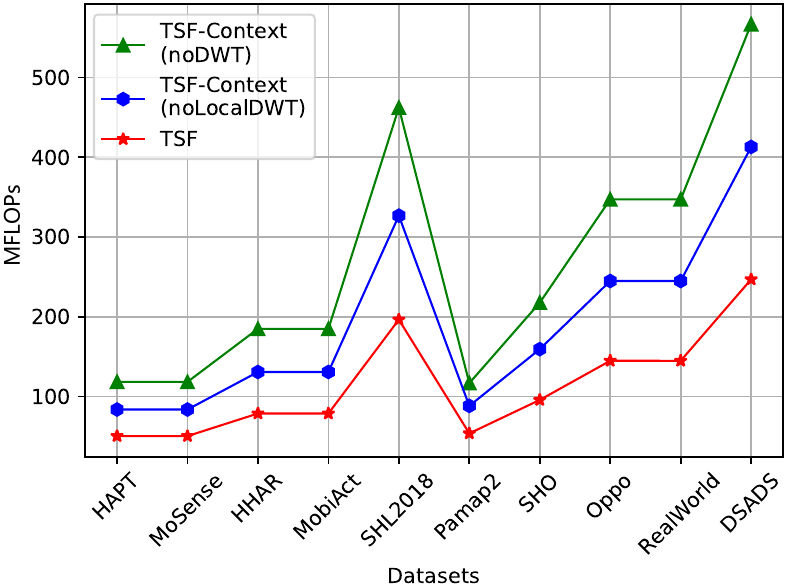}\\

\end{tabular}
\caption{Comparative analysis of FLOPs across multiple datasets for TSF, TSF (noLocalDWT), and TSF (noDWT) models.}
\label{fig6}
\end{figure}

Additionally, the FLOPs of TSF and its variants are presented in Figure~\ref{fig6}, showing TSF has significantly lower FLOPs than TSF-Context (noLocalDWT) and TSF-Context (noDWT). This indicates that removing the adaptive wavelet frequency selection from the temporal information fusion blocks results in a substantial increase in computational complexity. These findings demonstrate that adaptive frequency selection not only enhances recognition performance but also effectively reduces computational demands.

\textbf{(2) Impact of Selection of the Primary Frequency Component}

In the TSF-Context (Low-freq Prim) and TSF-Context (High-freq Prim) models, the primary wavelet frequency component within each temporal fusion block is set to low-frequency or high-frequency bands, respectively. The TSF-Context (Low-freq Prim) model shows a 0.50$\%$/0.37$\%$ average reduction in F1/WF1 scores across ten datasets (87.68$\%$/93.41$\%$ vs. 88.18$\%$/93.78$\%$), whereas the TSF-Context (High-freq Prim) model experiences a more substantial average decrease of 1.97$\%$/1.13$\%$ (86.21$\%$/92.65$\%$ vs. 88.18$\%$/93.78$\%$). The TSF-Context (Low-freq Prim) outperforms the high-frequency counterpart but still falls short of the original TSF model's performance. This is attributed to the prevalence of low-frequency components in most samples. Nonetheless, for samples with high-frequency activity, the TSF-Context (Low-freq Prim) model does not perform as well, unlike the TSF model which dynamically adjusts to both low-frequency and high-frequency activities, resulting in superior performance.

\textbf{(3) Replacing Adaptive Wavelet Frequency Selection with Pooling Operations}

In the TSF-Context (Pooling) model, the adaptive wavelet frequency selection is substituted with pooling operations. Table~\ref{tab7} shows that this change leads to an average performance drop of 1.57$\%$/1.35$\%$ in F1/WF1 scores (86.61$\%$/92.43$\%$ vs. 88.18$\%$/93.78$\%$) when compared to the standard TSF. Additionally, TSF-Context (Pooling) slightly underperforms TSF-Context (noDWT) (86.61$\%$/92.43$\%$ vs. 87.26$\%$/93.05$\%$). This is attributed to the pooling operations' fixed pattern of discarding timestamps, which results in substantial information loss due to lack data adaptability. In contrast, adaptive wavelet frequency selection retains critical context information, leading to better performance.

\begin{table*}[t]
	\centering
	\renewcommand\arraystretch{1.4}
    \scriptsize
	\caption{Results of ablation studies highlighting the impact of various modifications to the temporal information fusion block on F1 and WF1 scores across diverse datasets.}\label{tab7}
	\setlength{\tabcolsep}{0.68mm}{
	\begin{tabular}{cc|c|c|c|c|c|c|c|c|c}
		\midrule[0.75pt]
		\multirow{2}*{Methods} & {HAPT}  &  {MotionSense}  & {SHL2018} & {HHAR} & {MobiAct} & {Oppo} & {Pamap2} & {RealWorld} & {DSADS} & {SHO}\\
		\cmidrule(r){2-2} \cmidrule(r){3-3} \cmidrule(r){4-4} \cmidrule(r){5-5} \cmidrule(r){6-6} \cmidrule(r){7-7} \cmidrule(r){8-8} \cmidrule(r){9-9} \cmidrule(r){10-10} \cmidrule(r){11-11}
		&F1/WF1($\%$)     &F1/WF1($\%$)     &F1/WF1($\%$)     &F1/WF1($\%$)     &F1/WF1($\%$)     &F1/WF1($\%$)     &F1/WF1($\%$)     &F1/WF1($\%$)     &F1/WF1($\%$)       &F1/WF1($\%$) \\\midrule[0.75pt]
			\multirow{3}{*}{\textbf{\shortstack{TSF-Context \\ (noDWT)}}}
			&84.60/94.49          &96.35/96.93           &79.31/81.34           &94.63/94.82          &91.02/96.72          &\textbf{54.90}/89.46          &89.94/94.45          &94.74/95.16         &88.11      &99.07                \tabularnewline
            &$\pm$0.52/$\pm$0.20  &$\pm$0.14/$\pm$0.12   &$\pm$0.14/$\pm$0.23   &$\pm$0.14/$\pm$0.19  &$\pm$0.41/$\pm$0.08  &$\pm$0.55/$\pm$0.18  &$\pm$0.60/$\pm$0.26  &$\pm$0.31/$\pm$0.30 &$\pm$0.62  &$\pm$0.26            \tabularnewline
            &(0)/(0)              &(0)/(0)               &(0)/(0)               &(0)/(0)              &(0)/(0)              &(0)/(0)              &(0)/(0)              &(0)/(0)             &(0)        &(0)                  \tabularnewline
			\hline
            \multirow{3}{*}{\textbf{\shortstack{TSF-Context \\ (noLocal \\ DWT)}}}
			&85.14/\underline{94.56}          &96.70/97.26           &80.16/\underline{82.36}           &95.39/95.63          &91.90/96.68          &54.46/\underline{89.55}           &91.27/\underline{94.62}          &94.46/95.17         &88.57      &99.00                \tabularnewline
            &$\pm$0.34/$\pm$0.07  &$\pm$0.35/$\pm$0.31   &$\pm$0.77/$\pm$0.86   &$\pm$0.41/$\pm$0.45  &$\pm$0.25/$\pm$0.12  &$\pm$0.54/$\pm$0.08   &$\pm$0.37/$\pm$0.24  &$\pm$0.39/$\pm$0.44 &$\pm$0.59  &$\pm$0.08            \tabularnewline
            &(0.54)/(0.07)        &(0.35)/(0.33)         &(0.85)/(1.02)         &(0.76)/(0.81)        &(0.88)/(-0.04)       &(-0.44)/(0.09)        &(1.33)/(0.17)        &(-0.28)/(0.01)      &(0.46)     &(-0.07)               \tabularnewline
		    \hline
            \multirow{3}{*}{\textbf{\shortstack{TSF-Context \\ (noGlobal \\ DWT)}}}
			&\underline{85.35}/94.38          &96.64/97.18           &79.48/81.35           &94.68/94.90          &91.20/96.73          &\underline{54.89}/\textbf{89.60}           &\underline{91.55}/94.45          &94.71/\underline{95.51}         &88.73      &98.90                \tabularnewline
            &$\pm$0.22/$\pm$0.24  &$\pm$0.07/$\pm$0.02   &$\pm$0.24/$\pm$0.54   &$\pm$0.58/$\pm$0.60  &$\pm$0.21/$\pm$0.03  &$\pm$0.35/$\pm$0.17   &$\pm$0.42/$\pm$0.35  &$\pm$0.50/$\pm$0.05 &$\pm$0.21  &$\pm$0.34            \tabularnewline
            &(0.75)/(-0.11)       &(0.29)/(0.25)         &(0.17)/(0.01)         &(0.05)/(0.08)        &(0.18)/(0.01)        &(-0.01)/(0.14)        &(1.61)/(0.00)        &(-0.03)/(0.35)      &(0.62)     &(-0.17)               \tabularnewline
			\hline
            \multirow{3}{*}{\textbf{\shortstack{TSF-Context \\ (Low-freq \\ Prim)}}}
			&85.16/94.46          &96.98/\underline{97.44}           &\underline{80.59}/\underline{82.36}           &95.60/95.83          &\underline{92.17}/\underline{96.79}          &52.41/89.21          &90.70/94.42          &\underline{94.89}/95.30         &\underline{89.20}      &\underline{99.09}                \tabularnewline
    		&$\pm$0.40/$\pm$0.20  &$\pm$0.58/$\pm$0.45   &$\pm$0.28/$\pm$0.28   &$\pm$0.11/$\pm$0.12  &$\pm$0.27/$\pm$0.07  &$\pm$0.42/$\pm$0.15  &$\pm$0.76/$\pm$0.04  &$\pm$0.05/$\pm$0.09 &$\pm$0.77  &$\pm$0.04            \tabularnewline
            &(0.56)/(-0.03)       &(0.63)/(0.51)         &(0.28)/(1.02)         &(0.97)/(1.01)        &(1.15)/(0.07)        &(-2.49)/(-0.25)      &(0.76)/(-0.03)       &(0.15)/(0.14)       &(1.09)     &(0.02)              \tabularnewline
			\hline
            \multirow{3}{*}{\textbf{\shortstack{TSF-Context \\ (High-freq \\ Prim)}}}
			&84.53/94.19          &\underline{97.06}/97.34           &79.74/82.31           &\textbf{95.95}/\textbf{96.09}          &91.92/96.78          &44.11/87.09           &90.58/94.03          &94.28/94.73         &85.33      &98.60                \tabularnewline
            &$\pm$0.48/$\pm$0.23  &$\pm$0.20/$\pm$0.17   &$\pm$0.60/$\pm$0.49   &$\pm$0.23/$\pm$0.22  &$\pm$0.22/$\pm$0.05  &$\pm$0.78/$\pm$0.30   &$\pm$0.66/$\pm$0.04  &$\pm$0.21/$\pm$0.21 &$\pm$0.50  &$\pm$0.06            \tabularnewline
            &(-0.07)/(-0.30)      &(0.71)/(0.41)         &(0.43)/(0.97)         &(1.32)/(1.26)        &(0.90)/(0.06)        &(-10.79)/(2.37)       &(0.64)/(-0.42)      &(-0.46)/(-0.43)     &(-2.78)    &(-0.47)               \tabularnewline
			\hline
            \multirow{3}{*}{\textbf{\shortstack{TSF-Context \\ (Pooling)}}}
			&83.95/94.17	      &94.89/95.13	         &77.62/79.49	        &94.03/94.20	          &90.38/96.24	        &52.42/89.02	       &90.99/93.77	     &94.71/95.23	       &88.29	    &98.80            \tabularnewline
            &$\pm$0.43/$\pm$0.24  &$\pm$0.24/$\pm$0.28	 &$\pm$0.76/$\pm$1.19	&$\pm$0.42/$\pm$0.46  &$\pm$0.49/$\pm$0.07	&$\pm$0.66/$\pm$0.17   &$\pm$0.63/$\pm$0.39	 &$\pm$0.31/$\pm$0.35	&$\pm$1.24	    &$\pm$0.52
            \tabularnewline
            &(-0.65)/(-0.32)	  &(-1.46)/(-1.80)       &(-1.69)/(-1.85)	    &(-0.60)/(-0.62)	      &(-0.64)/(-0.48)	    &(-2.48)/(-0.44)	   &(1.05)/(-0.68)	 &(-0.03)/(0.07)   &(0.18)	&(-0.27)    \tabularnewline
		    \hline
			\hline
            \multirow{3}{*}{\textbf{\shortstack{TSF}}}
			&\textbf{85.89}/\textbf{94.74}          &\textbf{97.34}/\textbf{97.82}           &\textbf{81.31}/\textbf{83.38}           &\underline{95.81}/\underline{96.02}          &\textbf{92.47}/\textbf{96.96}          &52.44/89.35           &\textbf{92.33}/\textbf{94.91}          &\textbf{95.11}/\textbf{95.60}         &\textbf{89.70}      &\textbf{99.35}                \tabularnewline
            &$\pm$0.56/$\pm$0.21  &$\pm$0.21/$\pm$0.18   &$\pm$0.16/$\pm$0.30    &$\pm$0.31/$\pm$0.24  &$\pm$0.37/$\pm$0.12 &$\pm$1.04/$\pm$0.31  &$\pm$0.81/$\pm$0.29  &$\pm$0.11/$\pm$0.10 &$\pm$1.15            &$\pm$0.07             \tabularnewline
            &(1.29)/(0.25)        &(0.99)/(0.89)         &(2.00)/(2.04)         &(1.18)/(1.20)        &(1.45)/(0.24)        &(-2.46)/(-0.11)       &(2.39)/(0.46)        &(0.37)/(0.44)       &(1.59)     &(0.28)               \tabularnewline
			\midrule[0.75pt]
	\end{tabular}}

\end{table*}

\section{Analysis}\label{sec:Analysis}

In this section, some interpretative analyses‌ of the IMU fusion block, the modality node fusion block, and the temporal information fusion block are provided.

\subsection{Adaptive IMU Fusion for Dynamic Noise Reduction}

\begin{figure}[!h]
\centering
\footnotesize
    \begin{minipage}[b]{0.5\textwidth}
        \centering
        \subfloat[Comparison of WF1 scores with and without IMU fusion in the presence of high-frequency noise (left) and low-frequency noise (right)]{
        \includegraphics[width=4.33cm]{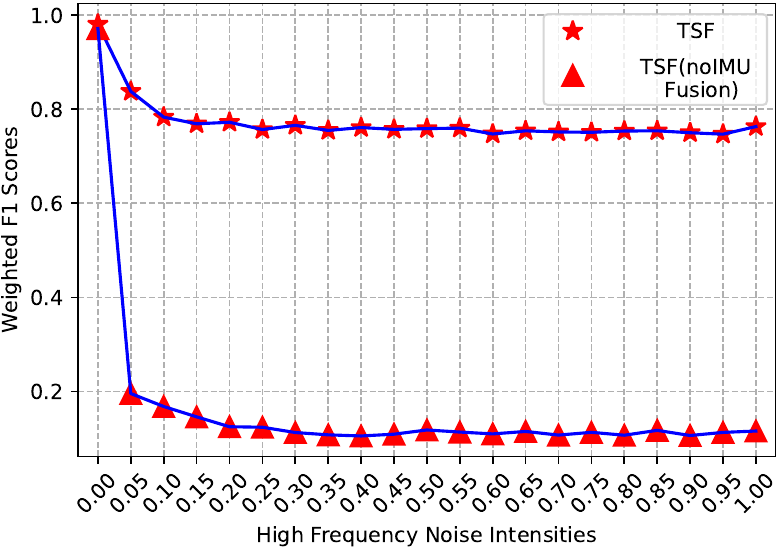}
        \includegraphics[width=4.33cm]{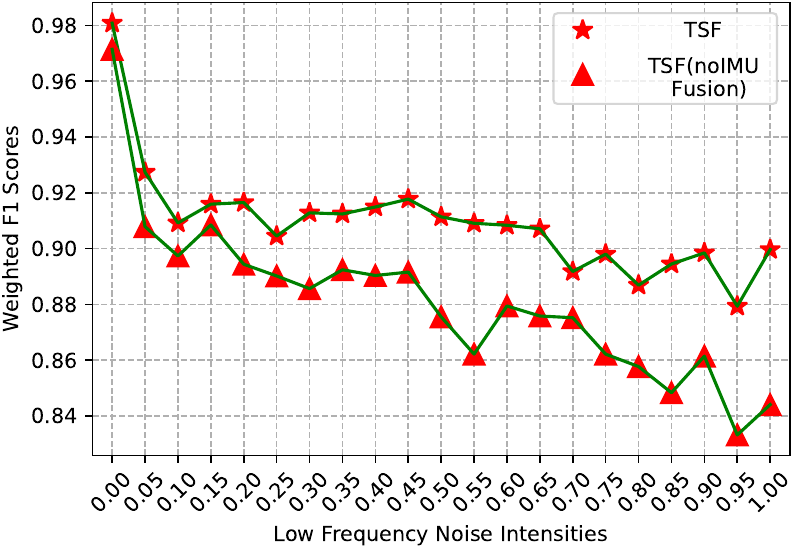}
        \label{fig7_a}
        }
    \vspace{0.2cm}
    \end{minipage}
    \begin{minipage}[b]{0.5\textwidth}
        \centering
        \subfloat[Average attention values for gravimeter features (left) and gyroscope features (right) across different levels of high-frequency and low-frequency noise intensities]{
        \includegraphics[width=4.33cm]{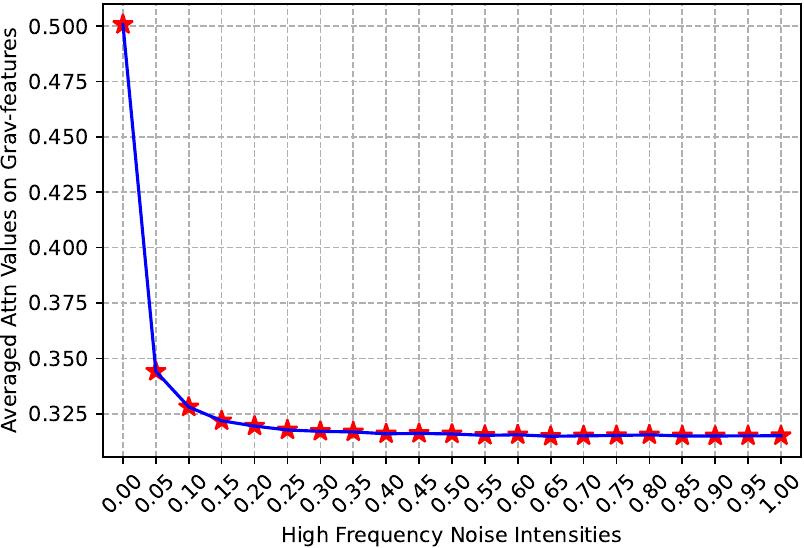}
        \includegraphics[width=4.33cm]{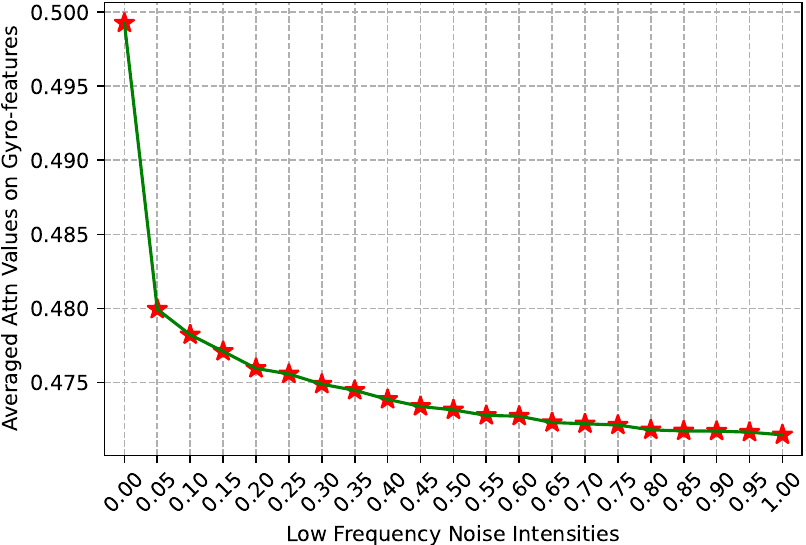}
        \label{fig7_b}
        }
    \end{minipage}
\caption{WF1 scores and mean posture-sensor attention values across various levels of high-frequency and low-frequency noise intensities, derived from samples of the first subject in the MotionSense dataset. Our IMU fusion block enhances noise robustness through adaptive attention weights.}
\label{fig7}
\end{figure}

Section~\ref{sec:4.2} explains that the IMU fusion block independently processes the gravimeter and gyroscope data through complementary low-pass and high-pass filters. This block uses the distinct frequency characteristics of gravimeters and gyroscopes for effective noise reduction in sensor fusion. To mimic real-world scenarios, varying levels of high-frequency and low-frequency noise are added separately to the gravimeter and gyroscope data. The WF1 scores of models with and without the IMU fusion block are compared across different noise levels (refer to Figure~\ref{fig7_a}). The results reveal a slower decline in the WF1 scores of the original TSF model as noise intensity increases. Figure~\ref{fig7_b} further illustrates this by showing average sensor attention values (refer to Equations~\ref{e4.4}--\ref{e4.6}) under different noise intensities. Notably, with increased low-frequency noise in gyroscope data, the attention values for gyroscopes decrease. This is linked to the relationship between gyroscope attention values and the cut-off frequency, where higher low-frequency noise raises the cut-off frequency, allowing more gravimeter features to pass through, and in turn reducing the attention given to gyroscopes. Conversely, adding high-frequency noise to gravimeter data lowers the cut-off frequency, making gyroscopes more dominant in the sensor fusion process and reducing attention to gravimeters. The attention values tend to stabilize at higher noise intensities, indicating that an optimal cut-off frequency is achieved.

In conclusion, the IMU fusion block is capable of adapting to varying noise conditions, making it a valuable tool for handling the complex and dynamic sensor noise in real-world scenarios. Its ability to adjust the cut-off frequency plays a key role in enhancing IMU sensor fusion.

\subsection{Graph Fourier Domain Filtering for Node Information Fusion in Diverse Activities}

Section~\ref{sec:4.3} focuses on applying adaptive filtering in the graph Fourier domain to merge diverse node information, both homogeneous and heterogeneous. The sign of edge values in the network guides this process: negative for heterogeneous and positive for homogeneous node information. This subsection categorizes graph nodes into two types: posture and motion. Edges connecting nodes within the same category but different body positions are termed as intra-edges, while those linking nodes from different categories are called inter-edges. By analyzing the weight values of these intra and inter-edges, the inclination of the modality node fusion block towards either homogeneous or heterogeneous node information extraction is discerned.

\begin{figure}[t]
	\centering
	\footnotesize
	\begin{minipage}[b]{0.5\textwidth}
        \centering
		\subfloat[`Biking', HHAR]{
			\label{fig8_a}
			\includegraphics[width=4.33cm]{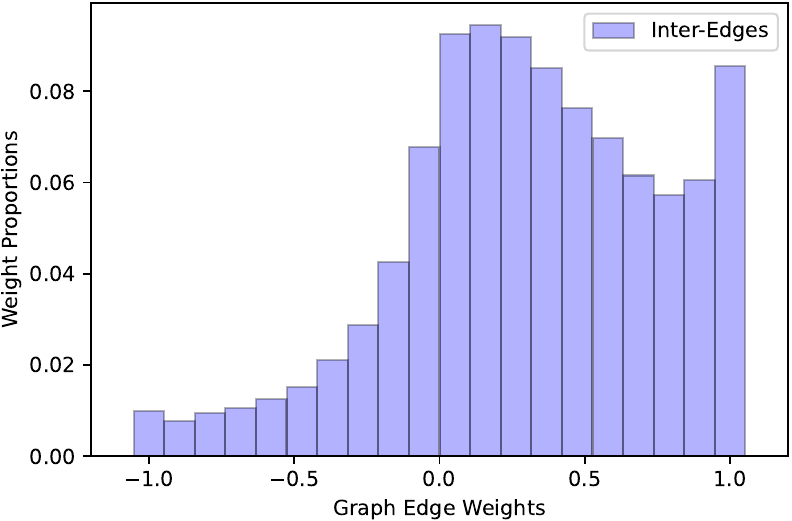}	
		}
		\subfloat[`Running on Treadmill', DSADS]{
			\label{fig8_b}
			\includegraphics[width=4.33cm]{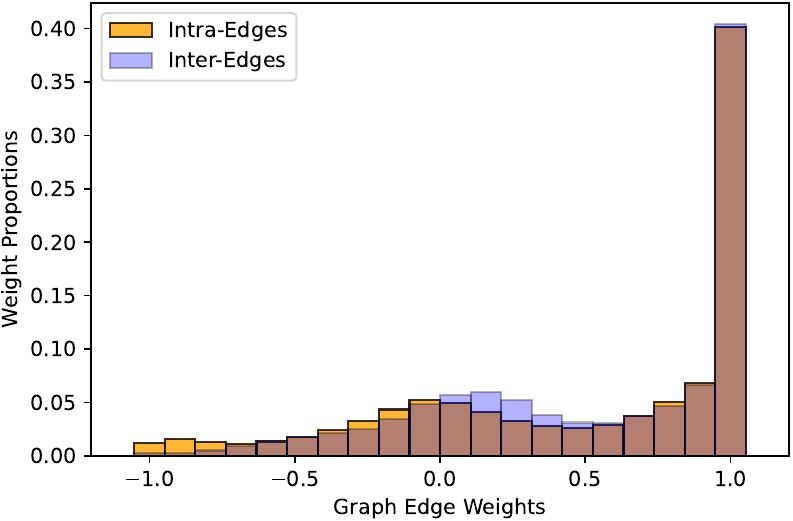}
		}
    \vspace{-0.1cm}
	\end{minipage}
	\begin{minipage}[b]{0.5\textwidth}
        \centering
		\subfloat[`Taking Subway', SHL2018]{
			\label{fig8_c}
			\includegraphics[width=4.33cm]{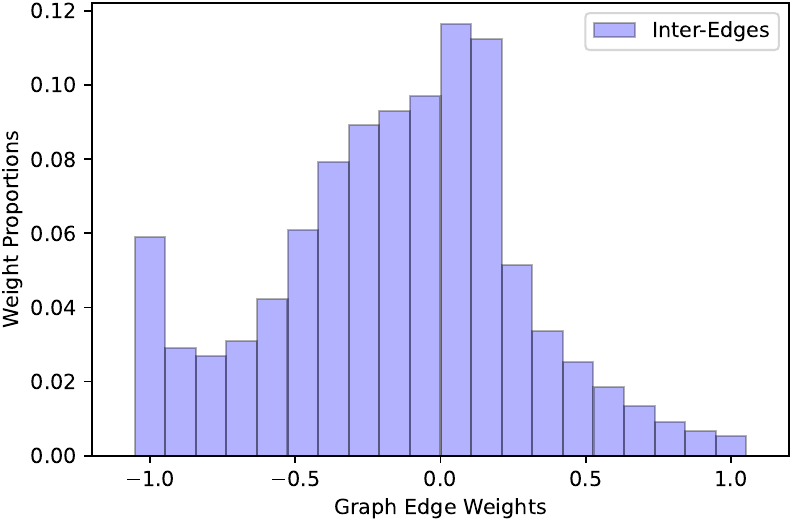}
		}
		\subfloat[`Playing Basketball', DSADS]{
			\label{fig8_d}
			\includegraphics[width=4.33cm]{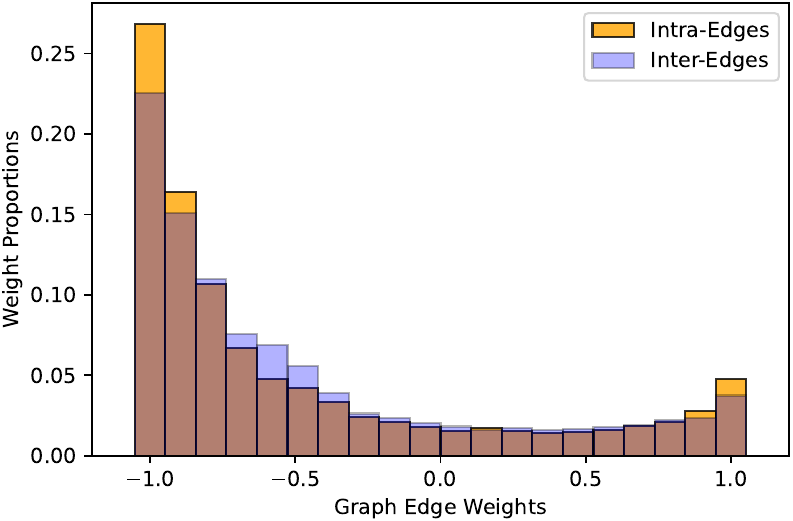}
		}
    \vspace{-0.1cm}
	\end{minipage}
	\begin{minipage}[b]{0.5\textwidth}
        \centering
		\subfloat[`Sitting', PAMAP2]{
			\label{fig8_e}
			\includegraphics[width=4.33cm]{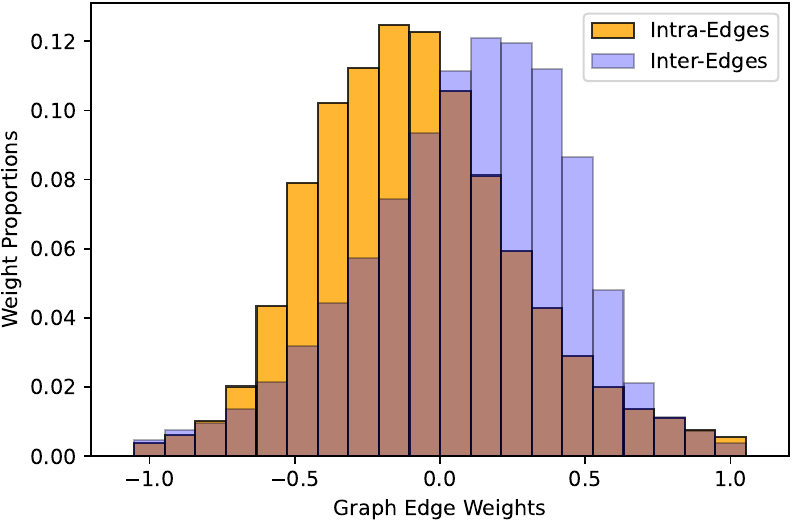}
		}
		\subfloat[`Closing Drawer', Oppo]{
			\label{fig8_f}
			\includegraphics[width=4.33cm]{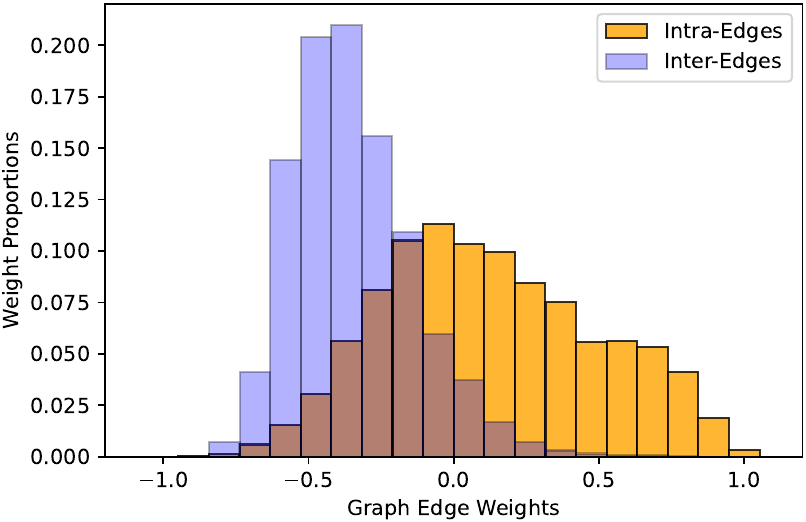}
		}
	\end{minipage}
	\caption{The histograms of intra-edge and inter-edge graph weights for different activities across multiple datasets. Since the HHAR and SHL2018 datasets are recorded by smartphones with a single IMU, these two datasets do not contain intra-edges. As shown, the learned graph weights adapt to different activity scenarios effectively.}
	\label{fig8}
\end{figure}

Figure~\ref{fig8} presents six histograms that detail edge weights for various activities across multiple datasets. Activities depicted in Figures~\ref{fig8_a} and \ref{fig8_b} involve specific equipments, like a bike or a treadmill, requiring subjects to follow fixed movement patterns. This results in strong correlations among graph nodes, leading the network to primarily extract homogeneous node information, as indicated by the dominance of positive weights in histograms. Conversely, activities in Figures~\ref{fig8_c} (`Taking Subway') and \ref{fig8_d} (`Playing Basketball') are less restricted, allowing for a wider range of posture and motion states, leading to weaker correlations between graph nodes. In these cases, the unique information of specific nodes becomes more significant (for instance, motion node information is key in identifying the `Taking Subway' activity). Consequently, the network is inclined to extract heterogeneous node information, as reflected in the histograms being mostly filled with negative weights.

Moreover, certain instances exhibit varying histogram distributions for intra- and inter-edges. In Figure~\ref{fig8_e}, for example, intra-edges mostly show negative weights, while inter-edges display predominantly positive weights. This pattern is observed in the PAMAP2 dataset, which uses three IMUs placed at various body locations. The `Sitting' activity, being relatively static, demonstrates strong links between posture and motion nodes, leading to a dominance of homogeneous information (mostly positive weights) among inter-edges. Conversely, to differentiate `Sitting' from other static activities like `Lying', the unique information from dispersed body positions becomes crucial, resulting in a dominance of heterogeneous information (mostly negative weights) among intra-edges. In contrast, the Oppo dataset depicted in Figure~\ref{fig8_f}, collected with seven wearables in close proximity, shows a different pattern during the `Closing Drawer' activity. Here, various body parts are closely interconnected, causing intra-edges to predominantly have positive weights. However, to distinguish `Closing Drawer' from similar activities like `Closing Door', the unique information from both posture and motion nodes is vital, leading to mainly negative weights in inter-edges.

Through adaptive filtering in the graph Fourier domain, our multi-modal fusion block adaptively merges both homogeneous and heterogeneous node information based on different activities and scenarios, thereby effectively executing the fusion of various modalities and positions.

\subsection{The Temporal Information Fusion Block for Context Correlation on Primary Wavelet Component}

Section~\ref{sec:4.4} explains how adaptive frequency selection layers select primary wavelet components tailored to the characteristics of each sample. This adaptability greatly reduces the length of the extracted features, enhancing the temporal information fusion for HAR. The time-frequency variations in recorded samples stem from three main sources: distinct activity habits of different participants, significant differences among various activities, and noticeable variations in repeated trials of the same activity by a single subject.

\begin{figure}[t]
\centering
\footnotesize
    \begin{minipage}[b]{0.5\textwidth}
        \centering
        \subfloat[Averaged spectral curves (left), temporal signals (right), and assigned DWT routes (legend part) of the data recorded by gyroscopes (y-axis is selected, which has obvious vibrations) on the waists of two subjects (sub$\_$1 and sub$\_$10 of the RealWorld dataset) performing the same activity (`Climbingup')]{
        \includegraphics[width=4.30cm]{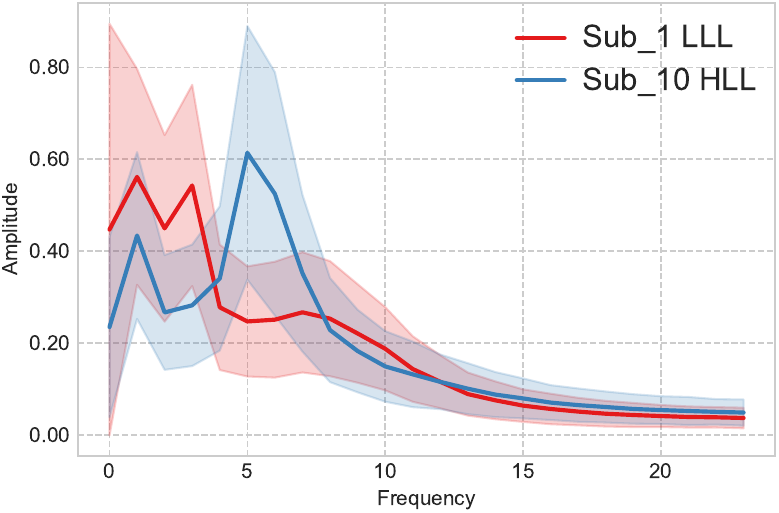}
        \includegraphics[width=4.33cm]{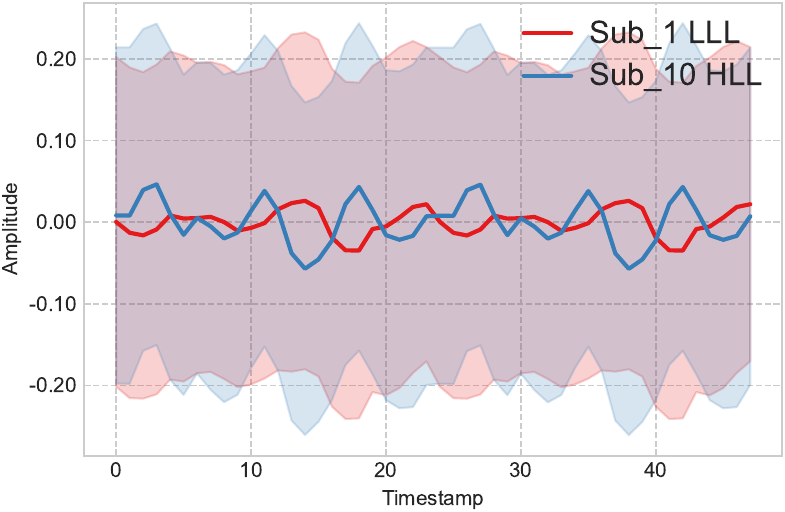}
        \label{fig9_a}
        }
    \vspace{0.2cm}
    \end{minipage}
    \begin{minipage}[b]{0.5\textwidth}
        \centering
        \subfloat[Averaged spectral curves (left), temporal signals (right), and assigned DWT routes (legend part) of the data recorded by the gyroscope (z-axis is selected) on a subject performing two different activities (`Stilling' and `Biking' of the SHL2018 dataset)]{
        \includegraphics[width=4.30cm]{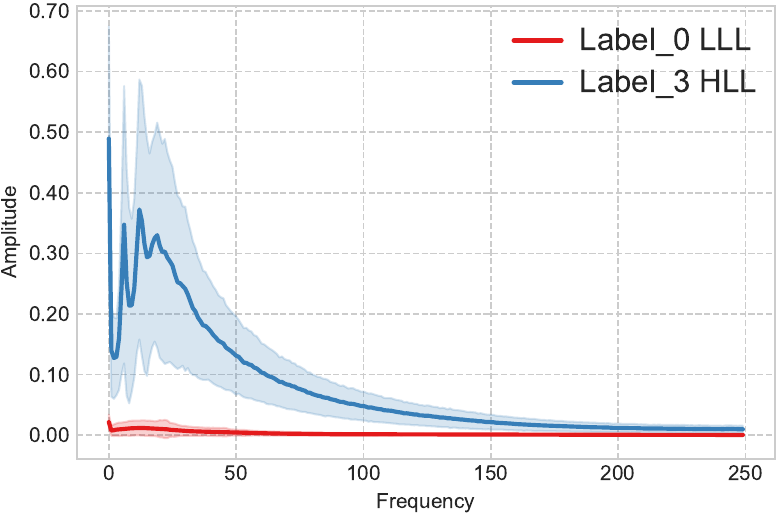}
        \includegraphics[width=4.33cm]{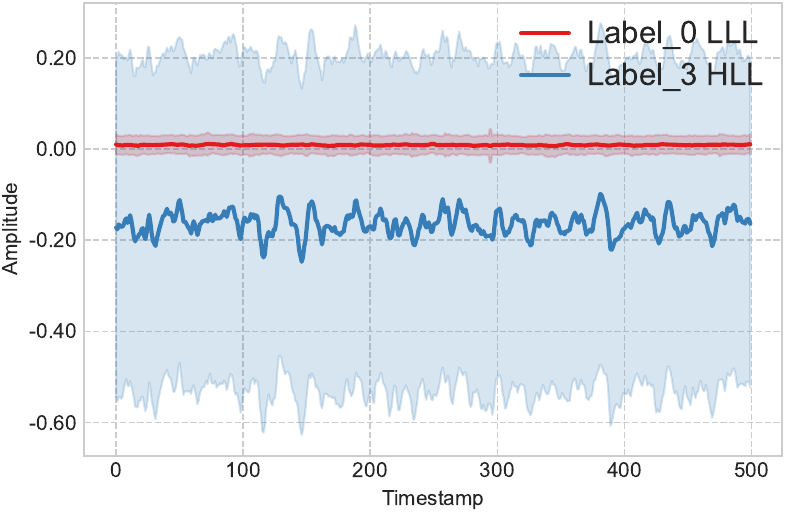}
        \label{fig9_b}
        }
    \vspace{0.2cm}
    \end{minipage}
    \begin{minipage}[b]{0.5\textwidth}
        \centering
        \subfloat[Averaged spectral curves (left), temporal signals (right), and assigned DWT routes (legend part) of the data recorded by the gyroscope (y-axis is selected) on the right leg of a subject performing the same activity (`Moving in Elevator' of the DSADS dataset) at multiple trials]{
        \includegraphics[width=4.30cm]{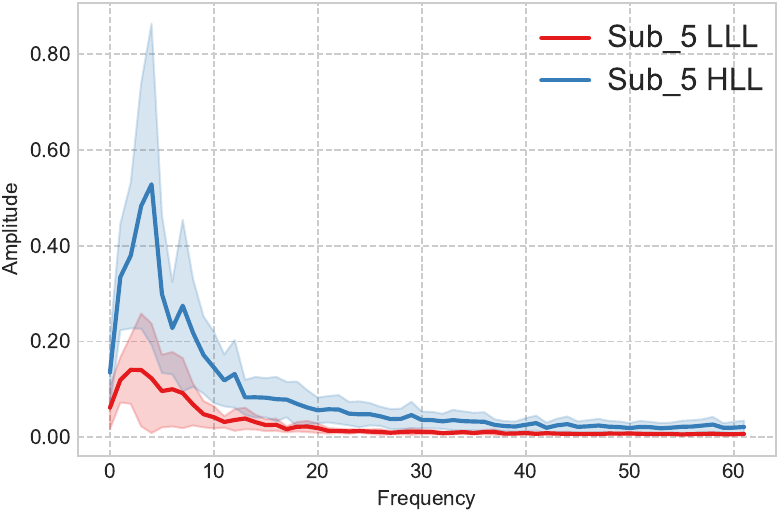}
        \includegraphics[width=4.33cm]{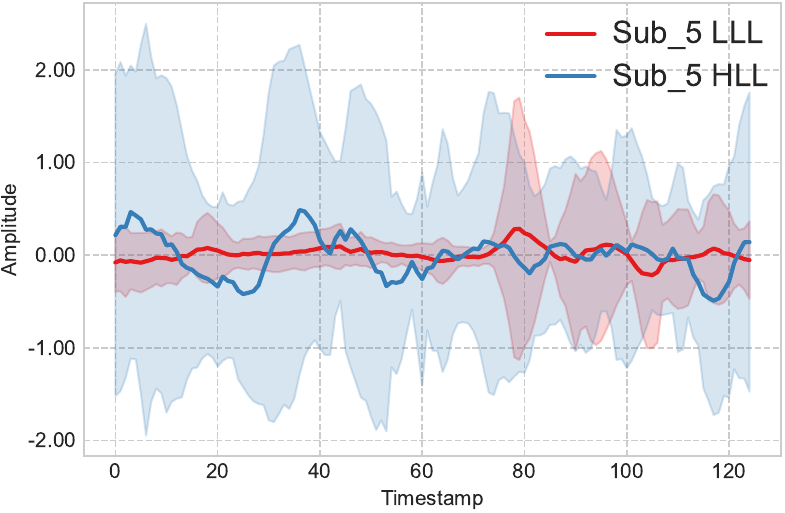}
        \label{fig9_c}
        }
    \end{minipage}
\caption{Averaged Fourier spectral curves and temporal signals of the recorded data corresponding to different DWT routes. L and H of the routes indicate separately selecting the low-frequency or high-frequency band as the primary component at each wavelet decomposition level (totally three times of decomposition are performed). The shaded parts represent the data distribution of corresponding samples. As observed, different samples are assigned to adaptive DWT routes according to their frequency characteristics.}
\label{fig9}
\end{figure}

Figure~\ref{fig9} provides examples of how adaptive wavelet decomposition is applied under these three scenarios. For instance, in the RealWorld dataset, Subject 10 climbs faster than Subject 1 (refer to Figure~\ref{fig9_a}). In the SHL2018 dataset, the frequency of fluctuations in the `Biking' activity is much higher than that in the `Sitting' activity (refer to Figure~\ref{fig9_b}). Even when the same participant in the DSADS dataset repeats the `Moving in Elevator' activity, the speed can vary significantly across different trials (refer to Figure~\ref{fig9_c}). In response to these scenarios, the adaptive frequency selection layers create dynamic DWT routes specific to each subject, activity, and trial. These routes align with the time-frequency characteristics of the samples: signals from faster actions are assigned routes corresponding to higher frequencies, and slower actions are linked with lower frequencies (refer to the legend parts of Figure~\ref{fig9}).

In conclusion, this customization ensures that the frequency selection layers can effectively adapt to the complex time-frequency characteristics of different samples. Fixed wavelet decomposition routes, in contrast, would fail to accurately represent these diverse and complex activities.

\section{Conclusion} \label{sec:Conclusion}

This paper introduces the TSF framework, a novel HAR system. TSF is structured around three key spectral fusion blocks. The first, an IMU fusion block, employs adaptive complementary filtering to suppress noise and integrates sensors from each IMU into posture and motion nodes. Next is the modality node fusion block, which executes the fusion of both homogeneous and heterogeneous node information using adaptive filtering within the graph Fourier domain. The final component is the temporal information fusion block. This block uses adaptive wavelet frequency selection to diminish context redundancy and shorten feature length, thereby enhancing both timestamp-wise node aggregation and long-term context correlation. The TSF framework effectively combines filtering and fusion techniques across Fourier, graph Fourier, and wavelet domains to achieve efficient multi-sensor fusion and context correlation. Comprehensive testing across ten datasets has proven the TSF framework's exceptional performance in HAR tasks.

\bibliographystyle{IEEEtran}

\bibliography{TSF}

@InProceedings{ha2015multi,
  author    = {Ha, Sojeong and Yun, Jeong-Min and Choi, Seungjin},
  title     = {Multi-modal convolutional neural networks for activity recognition},
  booktitle = {International Conference on Systems, Man, and Cybernetics},
  year      = {2015},
  pages     = {3017--3022},
}

@InProceedings{jiang2015human,
  author    = {Jiang, Wenchao and Yin, Zhaozheng},
  title     = {Human activity recognition using wearable sensors by deep convolutional neural networks},
  booktitle = {Proceedings of the ACM International Conference on Multimedia},
  year      = {2015},
  pages     = {1307--1310},
}

@InProceedings{guo2016wearable,
  author    = {Guo, Haodong and Chen, Ling and Peng, Liangying and Chen, Gencai},
  title     = {Wearable sensor based multimodal human activity recognition exploiting the diversity of classifier ensemble},
  booktitle = {Proceedings of the ACM International Joint Conference on Pervasive and Ubiquitous Computing},
  year      = {2016},
  pages     = {1112--1123},
}

@Article{khan2017detecting,
  author    = {Khan, Shehroz S and Taati, Babak},
  title     = {Detecting unseen falls from wearable devices using channel-wise ensemble of autoencoders},
  journal   = {Expert Systems with Applications},
  year      = {2017},
  volume    = {87},
  pages     = {280--290},
  publisher = {Elsevier},
}

@Article{ordonez2016deep,
  author    = {Ord{\'o}{\~n}ez, Francisco Javier and Roggen, Daniel},
  title     = {Deep convolutional and {LSTM} recurrent neural networks for multimodal wearable activity recognition},
  journal   = {Sensors},
  year      = {2016},
  volume    = {16},
  number    = {1},
  pages     = {115--140},
  publisher = {Multidisciplinary Digital Publishing Institute},
}

@InProceedings{yao2017deepsense,
  author    = {Yao, Shuochao and Hu, Shaohan and Zhao, Yiran and Zhang, Aston and Abdelzaher, Tarek},
  title     = {Deepsense: A unified deep learning framework for time-series mobile sensing data processing},
  booktitle = {International Conference on World Wide Web},
  year      = {2017},
  pages     = {351--360},
}

@InProceedings{yang2015deep,
  author    = {Yang, Jianbo and Nguyen, Minh Nhut and San, Phyo Phyo and Li, Xiaoli and Krishnaswamy, Shonali},
  title     = {Deep convolutional neural networks on multi-channel time series for human activity recognition.},
  booktitle = {International Joint Conference on Artificial Intelligence},
  year      = {2015},
  volume    = {15},
  pages     = {3995--4001},
}

@InProceedings{laput2019sensing,
  author    = {Laput, Gierad and Harrison, Chris},
  title     = {Sensing fine-grained hand activity with smartwatches},
  booktitle = {Proceedings of the CHI Conference on Human Factors in Computing Systems},
  year      = {2019},
  pages     = {1--13},
}

@InProceedings{betancourt2020self,
  author    = {Betancourt, Carlos and Chen, Wen-Hui and Kuan, Chi-Wei},
  title     = {Self-attention networks for human activity recognition using wearable devices},
  booktitle = {International Conference on Systems, Man, and Cybernetics},
  year      = {2020},
  pages     = {1194--1199},
}

@Article{abedin2021attend,
  author    = {Abedin, Alireza and Ehsanpour, Mahsa and Shi, Qinfeng and Rezatofighi, Hamid and Ranasinghe, Damith C},
  title     = {Attend and Discriminate: Beyond the State-of-the-Art for Human Activity Recognition Using Wearable Sensors},
  journal   = {Proceedings of the ACM on Interactive, Mobile, Wearable and Ubiquitous Technologies},
  year      = {2021},
  volume    = {5},
  number    = {1},
  pages     = {1--22},
  publisher = {ACM New York, NY, USA},
}

@Article{liu2020globalfusion,
  author    = {Liu, Shengzhong and Yao, Shuochao and Li, Jinyang and Liu, Dongxin and Wang, Tianshi and Shao, Huajie and Abdelzaher, Tarek},
  title     = {GlobalFusion: A Global Attentional Deep Learning Framework for Multisensor Information Fusion},
  journal   = {Proceedings of the ACM on Interactive, Mobile, Wearable and Ubiquitous Technologies},
  year      = {2020},
  volume    = {4},
  number    = {1},
  pages     = {1--27},
  publisher = {ACM New York, NY, USA},
}

@Article{chen2021deep,
  author    = {Chen, Kaixuan and Zhang, Dalin and Yao, Lina and Guo, Bin and Yu, Zhiwen and Liu, Yunhao},
  title     = {Deep Learning for Sensor-based Human Activity Recognition: Overview, Challenges, and Opportunities},
  journal   = {ACM Computing Surveys},
  year      = {2021},
  volume    = {54},
  number    = {4},
  pages     = {1--40},
  publisher = {ACM New York, NY, USA},
}

@InProceedings{mahmud2020human,
  author    = {Mahmud, Saif and Tonmoy, M and Bhaumik, Kishor Kumar and Rahman, AKM and Amin, M Ashraful and Shoyaib, Mohammad and Khan, Muhammad Asif Hossain and Ali, Amin Ahsan},
  title     = {Human activity recognition from wearable sensor data using self-attention},
  booktitle = {European Conference on Artificial Intelligence},
  year      = {2020},
  pages     = {1332--1339},
}

@Article{parkka2006activity,
  author    = {Parkka, Juha and Ermes, Miikka and Korpipaa, Panu and Mantyjarvi, Jani and Peltola, Johannes and Korhonen, Ilkka},
  title     = {Activity classification using realistic data from wearable sensors},
  journal   = {IEEE Transactions on Information Technology in Biomedicine},
  year      = {2006},
  volume    = {10},
  number    = {1},
  pages     = {119--128},
  publisher = {IEEE},
}

@Article{altun2010comparative,
  author    = {Altun, Kerem and Barshan, Billur and Tun{\c{c}}el, Orkun},
  title     = {Comparative study on classifying human activities with miniature inertial and magnetic sensors},
  journal   = {Pattern Recognition},
  year      = {2010},
  volume    = {43},
  number    = {10},
  pages     = {3605--3620},
  publisher = {Elsevier},
}

@Article{ward2006activity,
  author    = {Ward, Jamie A and Lukowicz, Paul and Troster, Gerhard and Starner, Thad E},
  title     = {Activity recognition of assembly tasks using body-worn microphones and accelerometers},
  journal   = {IEEE Transactions on Pattern Analysis and Machine Intelligence},
  year      = {2006},
  volume    = {28},
  number    = {10},
  pages     = {1553--1567},
  publisher = {IEEE},
}

@Article{preece2008comparison,
  author    = {Preece, Stephen J and Goulermas, John Yannis and Kenney, Laurence PJ and Howard, David},
  title     = {A comparison of feature extraction methods for the classification of dynamic activities from accelerometer data},
  journal   = {IEEE Transactions on Biomedical Engineering},
  year      = {2008},
  volume    = {56},
  number    = {3},
  pages     = {871--879},
  publisher = {IEEE},
}

@Article{wu2014bayesian,
  author    = {Wu, Jiaxiang and Cheng, Jian},
  title     = {Bayesian co-boosting for multi-modal gesture recognition},
  journal   = {The Journal of Machine Learning Research},
  year      = {2014},
  volume    = {15},
  number    = {1},
  pages     = {3013--3036},
  publisher = {JMLR. org},
}

@InProceedings{fish2012feature,
  author    = {Fish, Benjamin and Khan, Ammar and Chehade, Nabil Hajj and Chien, Chieh and Pottie, Greg},
  title     = {Feature selection based on mutual information for human activity recognition},
  booktitle = {International Conference on Acoustics, Speech and Signal Processing},
  year      = {2012},
  pages     = {1729--1732},
}

@Article{mahony2008nonlinear,
  author    = {Mahony, Robert and Hamel, Tarek and Pflimlin, Jean-Michel},
  title     = {Nonlinear complementary filters on the special orthogonal group},
  journal   = {IEEE Transactions on Automatic Control},
  year      = {2008},
  volume    = {53},
  number    = {5},
  pages     = {1203--1218},
  publisher = {IEEE},
}

@Book{chung1997spectral,
  title     = {Spectral Graph Theory},
  publisher = {American Mathematical Soc.},
  year      = {1997},
  author    = {Chung, Fan RK and Graham, Fan Chung},
  number    = {92},
}

@InProceedings{kipf2016semi,
  author    = {Kipf, Thomas N and Welling, Max},
  title     = {Semi-supervised classification with graph convolutional networks},
  booktitle = {International Conference on Learning Representations},
  year      = {2016},
}

@Article{mallat1989theory,
  author    = {Mallat, Stephane G},
  title     = {A theory for multiresolution signal decomposition: the wavelet representation},
  journal   = {IEEE Transactions on Pattern Analysis and Machine Intelligence},
  year      = {1989},
  volume    = {11},
  number    = {7},
  pages     = {674--693},
  publisher = {IEEE},
}

@InProceedings{jung2007inertial,
  author    = {Jung, Dongwon and Tsiotras, Panagiotis},
  title     = {Inertial attitude and position reference system development for a small {UAV}},
  booktitle = {In AIAA Infotech@ Aerospace 2007 Conference and Exhibit},
  year      = {2007},
  pages     = {2763--2778},
}

@InProceedings{ma2019attnsense,
  author    = {Ma, Haojie and Li, Wenzhong and Zhang, Xiao and Gao, Songcheng and Lu, Sanglu},
  title     = {AttnSense: Multi-level Attention Mechanism For Multimodal Human Activity Recognition.},
  booktitle = {International Joint Conference on Artificial Intelligence},
  year      = {2019},
  pages     = {3109--3115},
}

@Article{wang2022survey,
  author    = {Wang, Xiao and Bo, Deyu and Shi, Chuan and Fan, Shaohua and Ye, Yanfang and Philip, S Yu},
  title     = {A survey on heterogeneous graph embedding: methods, techniques, applications and sources},
  journal   = {IEEE Transactions on Big Data},
  year      = {2022},
  publisher = {IEEE},
}

@Article{yadav2021review,
  author    = {Yadav, Santosh Kumar and Tiwari, Kamlesh and Pandey, Hari Mohan and Akbar, Shaik Ali},
  title     = {A review of multimodal human activity recognition with special emphasis on classification, applications, challenges and future directions},
  journal   = {Knowledge-Based Systems},
  year      = {2021},
  volume    = {223},
  pages     = {106970},
  publisher = {Elsevier},
}

@Article{jang2016categorical,
  author  = {Jang, Eric and Gu, Shixiang and Poole, Ben},
  title   = {Categorical reparameterization with gumbel-softmax},
  journal = {arXiv preprint arXiv:1611.01144},
  year    = {2016},
}

@Article{van2013separating,
  author    = {Van Hees, Vincent T and Gorzelniak, Lukas and Dean Le{\'o}n, Emmanuel Carlos and Eder, Martin and Pias, Marcelo and Taherian, Salman and Ekelund, Ulf and Renstr{\"o}m, Frida and Franks, Paul W and Horsch, Alexander and others},
  title     = {Separating movement and gravity components in an acceleration signal and implications for the assessment of human daily physical activity},
  journal   = {PloS One},
  year      = {2013},
  volume    = {8},
  number    = {4},
  pages     = {e61691},
  publisher = {Public Library of Science San Francisco, USA},
}

@Article{vaswani2017attention,
  author  = {Vaswani, Ashish and Shazeer, Noam and Parmar, Niki and Uszkoreit, Jakob and Jones, Llion and Gomez, Aidan N and Kaiser, {\L}ukasz and Polosukhin, Illia},
  title   = {Attention is all you need},
  journal = {Advances in Neural Information Processing Systems},
  year    = {2017},
  volume  = {30},
}

@Article{gjoreski2018university,
  author    = {Gjoreski, Hristijan and Ciliberto, Mathias and Wang, Lin and Morales, Francisco Javier Ordonez and Mekki, Sami and Valentin, Stefan and Roggen, Daniel},
  title     = {The university of {Sussex-Huawei} locomotion and transportation dataset for multimodal analytics with mobile devices},
  journal   = {IEEE Access},
  year      = {2018},
  volume    = {6},
  pages     = {42592--42604},
  publisher = {IEEE},
}

@Article{reyes2016transition,
  author    = {Reyes-Ortiz, Jorge-L and Oneto, Luca and Sam{\`a}, Albert and Parra, Xavier and Anguita, Davide},
  title     = {Transition-aware human activity recognition using smartphones},
  journal   = {Neurocomputing},
  year      = {2016},
  volume    = {171},
  pages     = {754--767},
  publisher = {Elsevier},
}

@Article{malekzadeh2020privacy,
  author    = {Malekzadeh, Mohammad and Clegg, Richard G and Cavallaro, Andrea and Haddadi, Hamed},
  title     = {Privacy and utility preserving sensor-data transformations},
  journal   = {Pervasive and Mobile Computing},
  year      = {2020},
  volume    = {63},
  pages     = {101132},
  publisher = {Elsevier},
}

@InProceedings{stisen2015smart,
  author    = {Stisen, Allan and Blunck, Henrik and Bhattacharya, Sourav and Prentow, Thor Siiger and Kj{\ae}rgaard, Mikkel Baun and Dey, Anind and Sonne, Tobias and Jensen, Mads M{\o}ller},
  title     = {Smart devices are different: Assessing and mitigatingmobile sensing heterogeneities for activity recognition},
  booktitle = {Proceedings of the ACM Conference on Embedded Networked Sensor Systems},
  year      = {2015},
  pages     = {127--140},
}

@InProceedings{chatzaki2016human,
  author       = {Chatzaki, Charikleia and Pediaditis, Matthew and Vavoulas, George and Tsiknakis, Manolis},
  title        = {Human daily activity and fall recognition using a smartphone’s acceleration sensor},
  booktitle    = {International Conference on Information and Communication Technologies for Ageing Well and E-Health},
  year         = {2016},
  pages        = {100--118},
  organization = {Springer},
}

@Article{chavarriaga2013opportunity,
  author    = {Chavarriaga, Ricardo and Sagha, Hesam and Calatroni, Alberto and Digumarti, Sundara Tejaswi and Tr{\"o}ster, Gerhard and Mill{\'a}n, Jos{\'e} del R and Roggen, Daniel},
  title     = {The Opportunity challenge: A benchmark database for on-body sensor-based activity recognition},
  journal   = {Pattern Recognition Letters},
  year      = {2013},
  volume    = {34},
  number    = {15},
  pages     = {2033--2042},
  publisher = {Elsevier},
}

@InProceedings{reiss2012introducing,
  author    = {Reiss, Attila and Stricker, Didier},
  title     = {Introducing a new benchmarked dataset for activity monitoring},
  booktitle = {International Symposium on Wearable Computers},
  year      = {2012},
  pages     = {108--109},
}

@Article{yao2018efficient,
  author    = {Yao, Rui and Lin, Guosheng and Shi, Qinfeng and Ranasinghe, Damith C},
  title     = {Efficient dense labelling of human activity sequences from wearables using fully convolutional networks},
  journal   = {Pattern Recognition},
  year      = {2018},
  volume    = {78},
  pages     = {252--266},
  publisher = {Elsevier},
}

@InProceedings{sztyler2016body,
  author    = {Sztyler, Timo and Stuckenschmidt, Heiner},
  title     = {On-body localization of wearable devices: An investigation of position-aware activity recognition},
  booktitle = {International Conference on Pervasive Computing and Communications},
  year      = {2016},
  pages     = {1--9},
}

@Article{shoaib2014fusion,
  author    = {Shoaib, Muhammad and Bosch, Stephan and Incel, Ozlem Durmaz and Scholten, Hans and Havinga, Paul JM},
  title     = {Fusion of smartphone motion sensors for physical activity recognition},
  journal   = {Sensors},
  year      = {2014},
  volume    = {14},
  number    = {6},
  pages     = {10146--10176},
  publisher = {MDPI},
}

@InProceedings{he2015delving,
  author    = {He, Kaiming and Zhang, Xiangyu and Ren, Shaoqing and Sun, Jian},
  title     = {Delving deep into rectifiers: Surpassing human-level performance on imagenet classification},
  booktitle = {International Conference on Computer Vision},
  year      = {2015},
  pages     = {1026--1034},
}

@Article{kingma2014adam,
  author  = {Kingma, Diederik P and Ba, Jimmy},
  title   = {Adam: A method for stochastic optimization},
  journal = {arXiv preprint arXiv:1412.6980},
  year    = {2014},
}

@InProceedings{zhang2017mixup,
  author    = {Zhang, Hongyi and Cisse, Moustapha and Dauphin, Yann N and Lopez-Paz, David},
  title     = {Mixup: Beyond empirical risk minimization},
  booktitle = {International Conference on Learning Representations},
  year      = {2018},
}

@Article{zhang2022if,
  author    = {Zhang, Ye and Wang, Longguang and Chen, Huiling and Tian, Aosheng and Zhou, Shilin and Guo, Yulan},
  title     = {{IF-ConvTransformer}: A Framework for Human Activity Recognition Using {IMU} Fusion and {ConvTransformer}},
  journal   = {Proceedings of the ACM on Interactive, Mobile, Wearable and Ubiquitous Technologies},
  year      = {2022},
  volume    = {6},
  number    = {2},
  pages     = {1--26},
  publisher = {ACM New York, NY, USA},
}

@InProceedings{Bai2018,
  author    = {Bai, Shaojie and Kolter, J Zico and Koltun, Vladlen},
  title     = {An empirical evaluation of generic convolutional and recurrent networks for sequence modeling},
  booktitle = {International Conference on Learning Representations Workshop},
  year      = {2018},
}

@Article{miao2022towards,
  author    = {Miao, Shenghuan and Chen, Ling and Hu, Rong and Luo, Yingsong},
  title     = {Towards a dynamic inter-sensor correlations learning framework for multi-sensor-based wearable human activity recognition},
  journal   = {Proceedings of the ACM on Interactive, Mobile, Wearable and Ubiquitous Technologies},
  year      = {2022},
  volume    = {6},
  number    = {3},
  pages     = {1--25},
  publisher = {ACM New York, NY, USA},
}

@InProceedings{li2021two,
  author    = {Li, Bing and Cui, Wei and Wang, Wei and Zhang, Le and Chen, Zhenghua and Wu, Min},
  title     = {Two-stream convolution augmented {Transformer} for human activity recognition},
  booktitle = {Proceedings of the AAAI Conference on Artificial Intelligence},
  year      = {2021},
  volume    = {35},
  number    = {1},
  pages     = {286--293},
}

@InProceedings{yang2022predictive,
  author    = {Yang, Hong and LaBella, Aidan and Desell, Travis},
  title     = {Predictive maintenance for general aviation using convolutional {Transformers}},
  booktitle = {Proceedings of the AAAI Conference on Artificial Intelligence},
  year      = {2022},
  volume    = {36},
  number    = {11},
  pages     = {12636--12642},
}

@Article{shao2023convboost,
  author    = {Shao, Shuai and Guan, Yu and Zhai, Bing and Missier, Paolo and Pl{\"o}tz, Thomas},
  title     = {ConvBoost: Boosting ConvNets for Sensor-based Activity Recognition},
  journal   = {Proceedings of the ACM on Interactive, Mobile, Wearable and Ubiquitous Technologies},
  year      = {2023},
  volume    = {7},
  number    = {2},
  pages     = {1--21},
  publisher = {ACM New York, NY, USA},
}

@Article{zhang2022multi,
  author    = {Zhang, Ye and Hou, Yi and OuYang, Kewei and Zhou, Shilin},
  title     = {Multi-scale signed recurrence plot based time series classification using inception architectural networks},
  journal   = {Pattern Recognition},
  year      = {2022},
  volume    = {123},
  pages     = {108385},
  publisher = {Elsevier},
}

@Article{ye2018learning,
  author    = {Ye, Jun and Qi, Guo-Jun and Zhuang, Naifan and Hu, Hao and Hua, Kien A},
  title     = {Learning compact features for human activity recognition via probabilistic first-take-all},
  journal   = {IEEE Transactions on Pattern Analysis and Machine Intelligence},
  year      = {2018},
  volume    = {42},
  number    = {1},
  pages     = {126--139},
  publisher = {IEEE},
}

@Article{li2023difFormer,
  author  = {Li, Bing and Cui, Wei and Zhang, Le and Zhu, Ce and Wang, Wei and Tsang, Ivor W. and Zhou, Joey Tianyi},
  title   = {DifFormer: Multi-Resolutional Differencing Transformer With Dynamic Ranging for Time Series Analysis},
  journal = {IEEE Transactions on Pattern Analysis and Machine Intelligence},
  year    = {2023},
  volume  = {45},
  number  = {11},
  pages   = {13586-13598},
  doi     = {10.1109/TPAMI.2023.3293516},
}

@Article{Eldele2023self,
  author  = {Eldele, Emadeldeen and Ragab, Mohamed and Chen, Zhenghua and Wu, Min and Kwoh, Chee-Keong and Li, Xiaoli and Guan, Cuntai},
  title   = {Self-Supervised Contrastive Representation Learning for Semi-Supervised Time-Series Classification},
  journal = {IEEE Transactions on Pattern Analysis and Machine Intelligence},
  year    = {2023},
  volume  = {45},
  number  = {12},
  pages   = {15604-15618},
  doi     = {10.1109/TPAMI.2023.3308189},
}

@Article{selesnick1998generalized,
  author    = {Selesnick, Ivan W and Burrus, C Sidney},
  title     = {Generalized digital Butterworth filter design},
  journal   = {IEEE Transactions on Signal Processing},
  year      = {1998},
  volume    = {46},
  number    = {6},
  pages     = {1688--1694},
  publisher = {IEEE},
}

@Article{ricotti2023wearable,
  author    = {Ricotti, Valeria and Kadirvelu, Balasundaram and Selby, Victoria and Festenstein, Richard and Mercuri, Eugenio and Voit, Thomas and Faisal, A Aldo},
  title     = {Wearable full-body motion tracking of activities of daily living predicts disease trajectory in Duchenne muscular dystrophy},
  journal   = {Nature Medicine},
  year      = {2023},
  volume    = {29},
  number    = {1},
  pages     = {95--103},
  publisher = {Nature Publishing Group US New York},
}

@Article{phan2022xsleepnet,
  author    = {Phan, Huy and Ch{\'e}n, Oliver Y and Tran, Minh C and Koch, Philipp and Mertins, Alfred and De Vos, Maarten},
  title     = {XSleepNet: Multi-view sequential model for automatic sleep staging},
  journal   = {IEEE Transactions on Pattern Analysis and Machine Intelligence},
  year      = {2022},
  volume    = {44},
  number    = {9},
  pages     = {5903--5915},
  publisher = {IEEE},
}

@article{haresamudram2025past,
  title={Past, present, and future of sensor-based human activity recognition using wearables: A surveying tutorial on a still challenging task},
  author={Haresamudram, Harish and Tang, Chi Ian and Suh, Sungho and Lukowicz, Paul and Ploetz, Thomas},
  journal={Proceedings of the ACM on Interactive, Mobile, Wearable and Ubiquitous Technologies},
  volume={9},
  number={2},
  pages={1--44},
  year={2025},
  publisher={ACM New York, NY, USA}
}

@article{thukral2025layout,
  title={Layout-agnostic human activity recognition in smart homes through textual descriptions of sensor triggers (tdost)},
  author={Thukral, Megha and Dhekane, Sourish Gunesh and Hiremath, Shruthi K and Haresamudram, Harish and Ploetz, Thomas},
  journal={Proceedings of the ACM on Interactive, Mobile, Wearable and Ubiquitous Technologies},
  volume={9},
  number={1},
  pages={1--38},
  year={2025},
  publisher={ACM New York, NY, USA}
}

@article{busso2025diversityone,
  title={DiversityOne: A multi-country smartphone sensor dataset for everyday life behavior modeling},
  author={Busso, Matteo and Bontempelli, Andrea and Malcotti, Leonardo Javier and Meegahapola, Lakmal and Kun, Peter and Diwakar, Shyam and Nutakki, Chaitanya and Britez, Marcelo Dario Rodas and Xu, Hao and Song, Donglei and others},
  journal={Proceedings of the ACM on Interactive, Mobile, Wearable and Ubiquitous Technologies},
  volume={9},
  number={1},
  pages={1--49},
  year={2025},
  publisher={ACM New York, NY, USA}
}

@inproceedings{xie2025decomposing,
  title={Decomposing and Fusing Intra- and Inter-Sensor Spatio-Temporal Signal for Multi-Sensor Wearable Human Activity Recognition},
  author={Xie, Haoyu and Li, Haoxuan and Zheng, Chunyuan and Yuan, Haonan and Liao, Guorui and Liao, Jun and Liu, Li},
  booktitle={Proceedings of the AAAI Conference on Artificial Intelligence},
  volume={39},
  number={13},
  pages={14441--14449},
  year={2025}
}

@article{chai2025iot,
  title={{IoT-FAR}: A multi-sensor fusion approach for {IoT}-based firefighting activity recognition},
  author={Chai, Xiaoqing and Lee, Boon Giin and Hu, Chenhang and Pike, Matthew and Chieng, David and Wu, Renjie and Chung, Wan-Young},
  journal={Information Fusion},
  volume={113},
  pages={102650},
  year={2025},
  publisher={Elsevier}
}

@article{li2025gesture,
  title={Gesture recognition with adaptive-weight-based residual MultiheadCrossAttention fusion based on multi-level feature information},
  author={Li, Zhuang and Shou, Dahua},
  journal={Information Fusion},
  volume={115},
  pages={102789},
  year={2025},
  publisher={Elsevier}
}

@inproceedings{wang2024multi,
  title={Multi-scale context-aware networks based on fragment association for human activity recognition},
  author={Wang, Zhiqiong and Liu, Hanyu and Zhao, Boyang and Shen, Qi and Li, Mingzhe and Que, Ningfeng and Yan, Mingke and Xin, Junchang},
  booktitle={International Joint Conference on Artificial Intelligence},
  pages={3169--3177},
  year={2024}
}

@article{jeong2024precyse,
  title={{PRECYSE}: Predicting cybersickness using transformer for multimodal time-series sensor data},
  author={Jeong, Dayoung and Han, Kyungsik},
  journal={Proceedings of the ACM on Interactive, Mobile, Wearable and Ubiquitous Technologies},
  volume={8},
  number={2},
  pages={1--24},
  year={2024},
  publisher={ACM New York, NY, USA}
}

@article{nguyen2025class,
  title={Class-Agnostic Repetitive Action Counting Using Wearable Devices},
  author={Nguyen, Duc Duy and Nguyen, Lam Thanh and Huang, Yifeng and Pham, Cuong and Hoai, Minh},
  journal={IEEE Transactions on Pattern Analysis and Machine Intelligence},
  volume={47},
  number={6},
  pages={4984--4995},
  year={2025},
  publisher={IEEE}
}

@article{dai2024hisc4d,
  title={{HiSC4D}: Human-centered interaction and {4D} scene capture in large-scale space using wearable {IMU}s and {LiDAR}},
  author={Dai, Yudi and Wang, Zhiyong and Lin, Xiping and Wen, Chenglu and Xu, Lan and Shen, Siqi and Ma, Yuexin and Wang, Cheng},
  journal={IEEE Transactions on Pattern Analysis and Machine Intelligence},
  volume={46},
  number={12},
  pages={11236--11253},
  year={2024},
  publisher={IEEE}
}

@article{yang2024cross,
  title={Cross-modal federated human activity recognition},
  author={Yang, Xiaoshan and Xiong, Baochen and Huang, Yi and Xu, Changsheng},
  journal={IEEE Transactions on Pattern Analysis and Machine Intelligence},
  volume={46},
  number={8},
  pages={5345--5361},
  year={2024},
  publisher={IEEE}
}

@article{li2025stade,
  title={{STADe}: Sensory Temporal Action Detection via Temporal-Spectral Representation Learning},
  author={Li, Bing and Duan, Haotian and Liu, Yun and Zhang, Le and Cui, Wei and Zhou, Joey Tianyi},
  journal={IEEE Transactions on Pattern Analysis and Machine Intelligence},
  volume={47},
  number={9},
  pages={8117--8133},
  year={2025},
  publisher={IEEE}
}

@article{wang2023exploring,
  title={Exploring fine-grained sparsity in convolutional neural networks for efficient inference},
  author={Wang, Longguang and Guo, Yulan and Dong, Xiaoyu and Wang, Yingqian and Ying, Xinyi and Lin, Zaiping and An, Wei},
  journal={IEEE Transactions on Pattern Analysis and Machine Intelligence},
  volume={45},
  number={4},
  pages={4474--4493},
  year={2023},
  publisher={IEEE}
}

@article{guo2026deep,
  title={Deep Lookup Network},
  author={Guo, Yulan and Wang, Longguang and Mao, Wendong and Dong, Xiaoyu and Wang, Yingqian and Liu, Li and An, Wei},
  journal={IEEE Transactions on Pattern Analysis and Machine Intelligence},
  volume={48},
  number={1},
  pages={202--218},
  year={2026},
  publisher={IEEE}
}

\begin{IEEEbiography}
[{\includegraphics[width=1in,height=1.25in,clip,keepaspectratio]{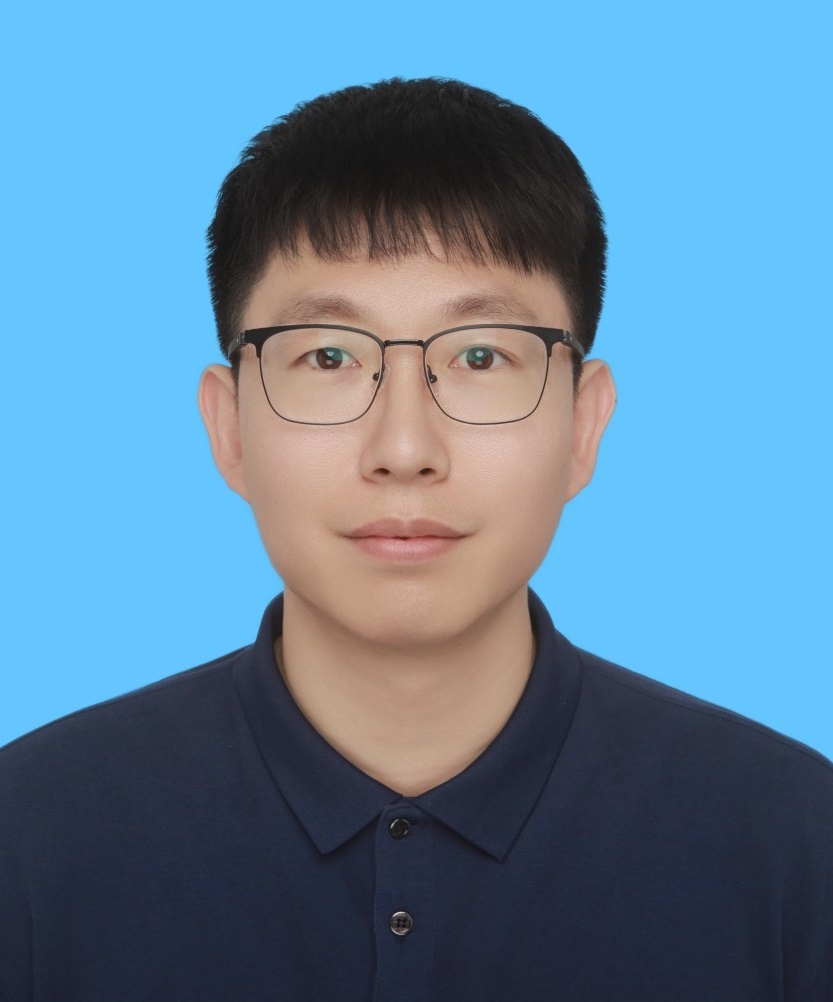}}]{Ye Zhang} received the Ph.D. degree in Electronic Science and Technology from National University of Defense Technology (NUDT), Changsha, China, in 2023. He served as a postdoctoral fellow at Sun Yat-sen University (SYSU) from 2023 to 2024. He is currently an associate researcher of the School of Electronics and Communication Engineering, Sun Yat-sen University, Shenzhen, China. His research interests include human-computer interaction and spatial intelligence.
\vspace{-1.6cm}
\end{IEEEbiography}

\begin{IEEEbiography}[{\includegraphics[width=1in,height=1.25in,clip,keepaspectratio]{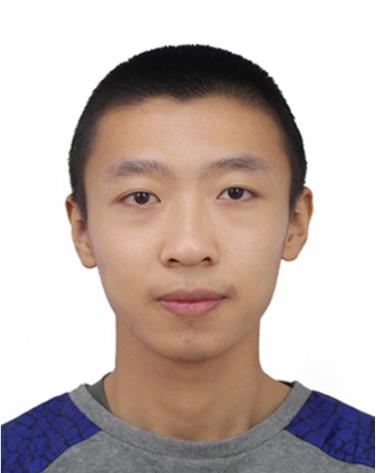}}]{Longguang Wang} received the B.E. degree in Electrical Engineering from Shandong University (SDU), Jinan, China, in 2015, and the Ph.D. degree in Information and Communication Engineering from National University of Defense Technology (NUDT), Changsha, China, in 2022. His current research interests include low-level vision and 3D vision.
\vspace{-1.6cm}
\end{IEEEbiography}

\begin{IEEEbiography}
[{\includegraphics[width=1in,height=1.25in,clip,keepaspectratio]{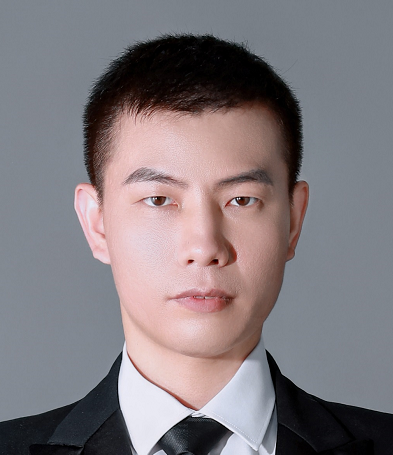}}]{Qing Gao} received the B.S. degree in automation from the School of Electrical Engineering and Automation, Liaoning University of Technology, Fuxin, China, in 2013, and the Ph.D. degree from the State Key Laboratory of Robotics, University of Chinese Academy of Sciences (CAS), Shenyang, China, in 2020. He is currently an Associate Professor with the School of Electronics and Communication Engineering, Sun Yat-sen University, Shenzhen, China. His research interests include robotics and human-robot interaction.
\vspace{-1.6cm}
\end{IEEEbiography}

\begin{IEEEbiography}
[{\includegraphics[width=1in,height=1.25in,clip,keepaspectratio]{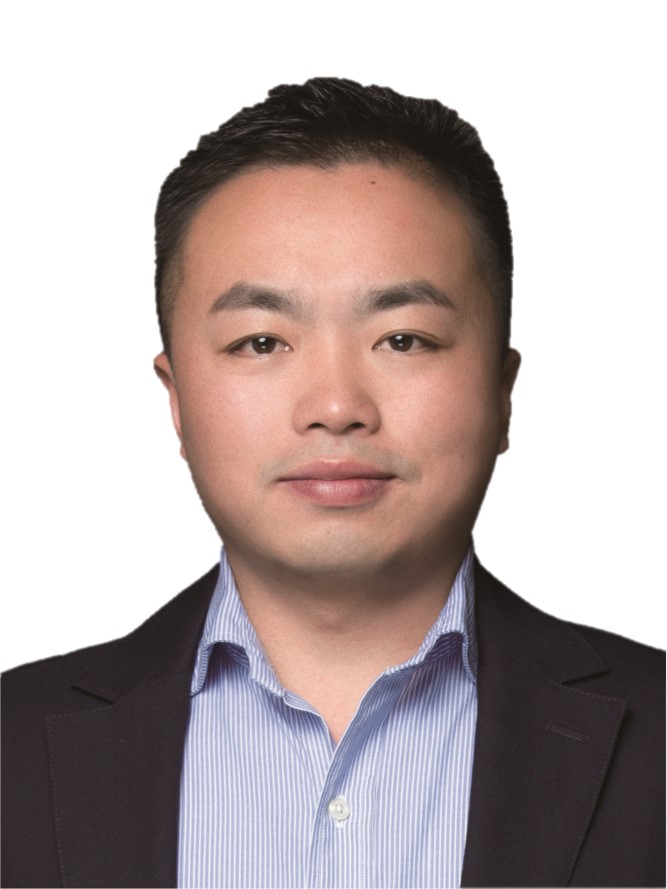}}]{Chaocan Xiang} (Member, IEEE) is a full professor at the College of Computer Science, Chongqing University, Chongqing, China. He received the BS and Ph.D. degrees in Computer Science and Engineering from the Nanjing Institute of Communication Engineering, China, in 2009 and 2014, respectively. He studied at the Real-Time Computing Lab, the University of Michigan, Ann Arbor in 2017. His current research interests include
crowd-sensing networks and IoT. He has published more than 30 research papers in important conferences and journals, such as ACM UbiComp, IEEE INFOCOM, IEEE/ACM TON, IEEE TMC, IEEE TPDS, IEEE T-ITS, IEEE TNSE, and
ACM TOSN.
\vspace{-1.6cm}
\end{IEEEbiography}

\begin{IEEEbiography}
[{\includegraphics[width=1in,height=1.25in,clip,keepaspectratio]{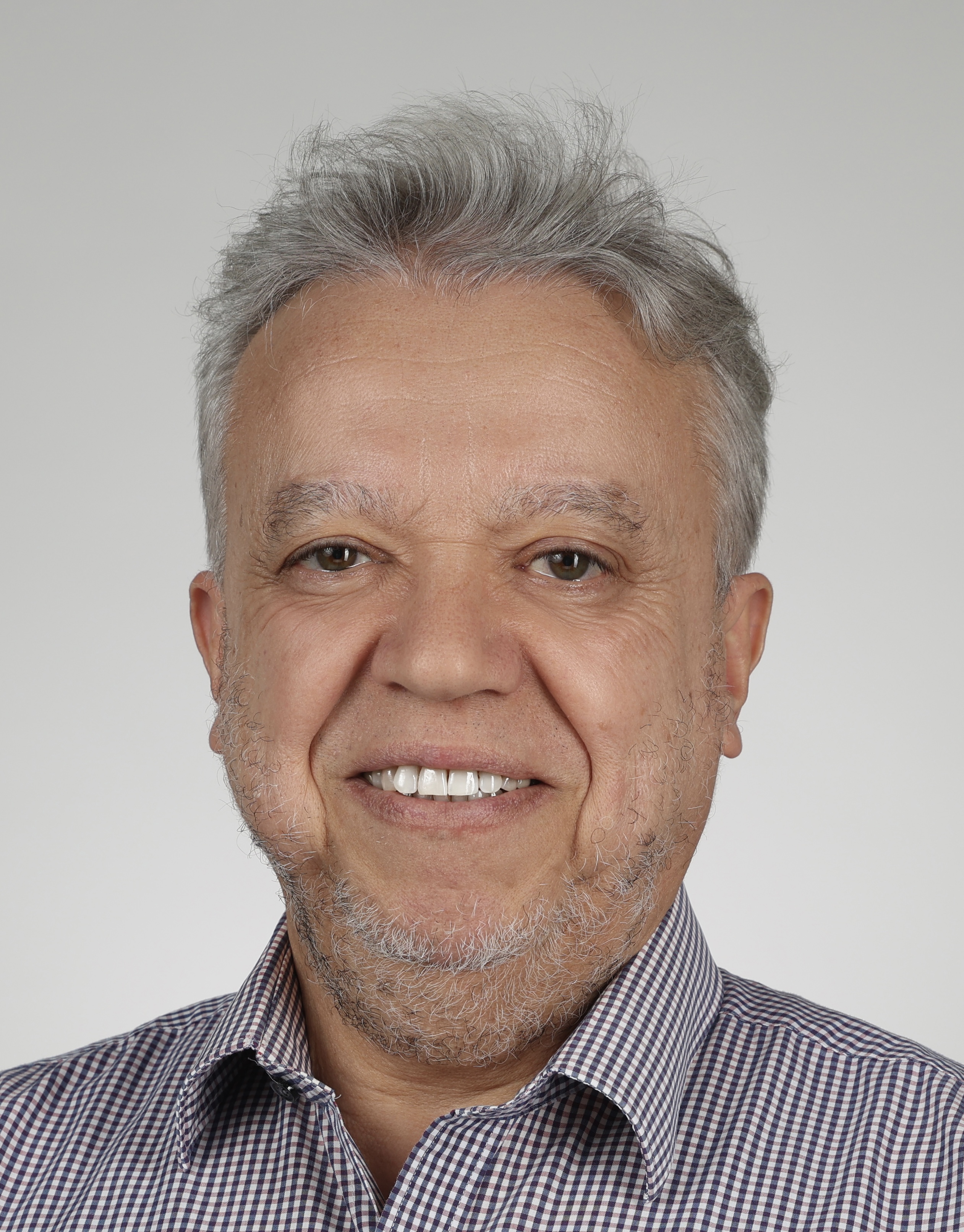}}]{Mohammed Bennamoun} is Winthrop Professor in the Department of Computer Science and Software Engineering at UWA and is a researcher in computer vision, machine/deep learning, robotics, and signal/speech processing. He has published 4 books (available on Amazon), 1 edited book, 1 Encyclopedia article, 14 book chapters, 200+ journal papers, 300+ conference publications, 16 invited keynote publications. His h-index is 87 and his number of citations is 42,000+ (Google Scholar). He was awarded 70+ competitive research grants, from the Australian Research Council, and numerous other Government, UWA and industry Research Grants.
\vspace{-1.6cm}
\end{IEEEbiography}

\begin{IEEEbiography}[{\includegraphics[width=1in,height=1.25in,clip,keepaspectratio]{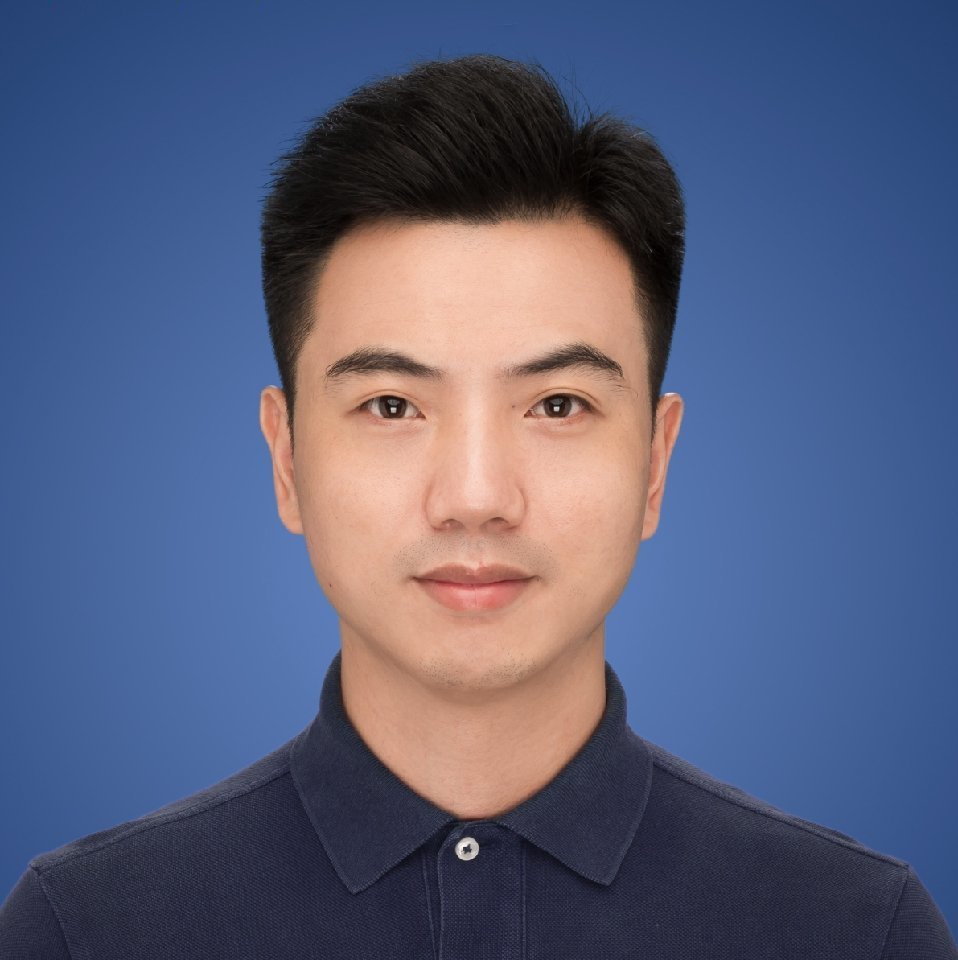}}]{Yulan Guo} is a full Professor with Sun Yat-sen University. He has authored over 200 articles at highly referred journals and conferences, receiving over 27,000 citations in Google Scholar. His research interests lie in spatial intelligence, 3D vision, and robotics. He served as a Senior Area Editor for IEEE Transactions on Image Processing, and an Associate Editor for the Visual Computer, and Computers \& Graphics. He also served as an area chair for CVPR, ICCV, ECCV, NeurIPS, and ICML. He organized over 10 workshops, challenges, and tutorials in prestigious conferences such as CVPR, ICCV, ECCV, and 3DV. He is a Senior Member of IEEE and ACM.
\end{IEEEbiography}

\end{document}